\documentclass[journal]{IEEEtran}
\usepackage{amsmath}
\usepackage{amssymb}

\DeclareMathOperator*{\argmin}{arg\,min}
\usepackage{hyperref}

\usepackage[
    natbib=true,
    style=numeric,
    sorting=none
]{biblatex}
\usepackage{dsfont}
\usepackage{graphicx}
\graphicspath{ {./images/} }

\usepackage{booktabs}
\usepackage{xcolor}  
\usepackage[capitalise]{cleveref}

\ifCLASSINFOpdf
\else
\fi

\usepackage{array}

\begin{document}

\title{Distance-Aware Occlusion Detection \\with Focused Attention}

\author{Yang~Li,
        Yucheng~Tu,
        Xiaoxue~Chen,
        Hao~Zhao,
        and~Guyue~Zhou
\thanks{This work was supported by Baidu and Anker. \textit{(Yang~Li and Yucheng~Tu contributed equally to this work.)} \textit{(Corresponding author: Guyue Zhou)}}
\thanks{Yang Li, Xiaoxue Chen, and Guyue Zhou are with the Institute for AI Industry Research, Tsinghua University, Beijing, China. (e-mail: yal009@ucsd.edu; chenxx21@mails.tsinghua.edu.cn; zhouguyue@air.tsinghua.edu.cn)}
\thanks{Yucheng Tu is with the Department
of Mathematics, University of California San Diego, La Jolla,
CA, 92093 USA (e-mail: y7tu@ucsd.edu).}
\thanks{Hao Zhao is with Intel Labs China and Peking University. (e-mail: hao.zhao@intel.com; zhao-hao@pku.edu.cn)}
\thanks{The source code is available at https://github.com/Yang-Li-2000/Distance-Aware-Occlusion-Detection-with-Focused-Attention.git.}
\thanks{This paper has supplementary demo videos available at https://youtu.be/gsfCxWO0xws and https://youtu.be/PfRZGlJXGOA provided by the authors.}
}

\markboth{Journal of \LaTeX\ Class Files,~Vol.~14, No.~8, August~2015}%
{Shell \MakeLowercase{\textit{et al.}}: Bare Demo of IEEEtran.cls for IEEE Journals}

\maketitle

      \begin{abstract}
For humans, understanding the relationships between objects using visual signals is intuitive. For artificial intelligence, however, this task remains challenging. Researchers have made significant progress studying \textcolor[RGB]{0,0,0}{semantic relationship detection}, such as human-object interaction detection and visual relationship detection. We take the study of visual relationships a step further from semantic to geometric. In specific, we predict relative occlusion and relative distance relationships. However, detecting these relationships from a single image is challenging. Enforcing focused attention to task-specific regions plays a critical role in successfully detecting these relationships. In this work, (1) we propose a novel three-decoder architecture as the infrastructure for focused attention; 2) we use the generalized intersection box prediction task to effectively guide our model to focus on occlusion-specific regions; 3) our model achieves a new state-of-the-art performance on distance-aware relationship detection. Specifically, our model increases the distance F1-score from 33.8\% to 38.6\% and boosts the occlusion F1-score from 34.4\% to  41.2\%. Our code is publicly available. 
\end{abstract}


\IEEEpeerreviewmaketitle

\section{Introduction}
\IEEEPARstart{V}isual object detection has shown exciting progress \citep{viola2004robust}\citep{girshick2015deformable} in the last two decades thanks to the representation learning power of tailored deep neural network architectures in 2D \citep{ren2016faster}\citep{cai2018cascade} and 3D \citep{qi2019deep}\citep{chen2021pq} settings. The community is now actively exploring the detection of higher-order semantic entities like human-object interactions (HOI)~\citep{li2017vip}\citep{gupta2019no}. 

In this paper, we go one step further from semantic to geometric. Specifically, we study two ubiquitous and essential relationships: relative distance and relative occlusion from the viewpoint. 

\begin{figure}[t]
    \centering
    \includegraphics[width=0.8\linewidth]{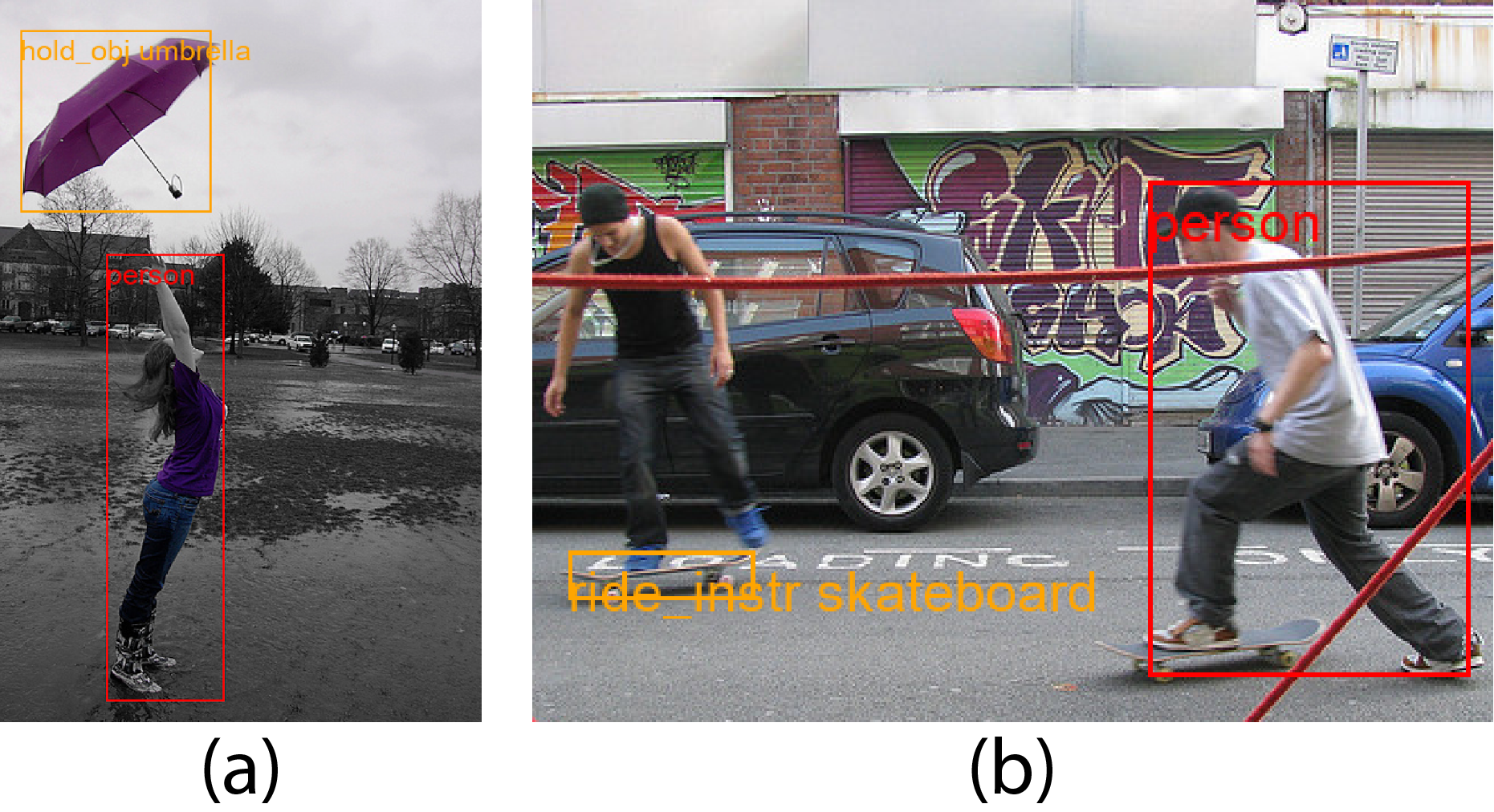}
    \caption{Understanding relative occlusion and relative distance may help HOI models rule out some unreasonable predictions such as (a) ``a woman is holding an umbrella" when her hand is not occluding the handle of the umbrella at all, and (b) ``a man (in the red box) is riding a skateboard (in the orange box)" that is much further to the viewpoint than the man.}
    \label{hoi_failures}
\end{figure}

\textcolor[RGB]{0,0,0}{Detecting these relationships from a 2D image is important because it will benefit other computer vision tasks, like HOI detection--the task of detecting human-object pairs and the relationships within each pair of them. Specifically, in HOI detection, models that explicitly possess the ability to understand relative occlusion and relative depth may more easily rule out some unreasonable predictions. For example, in \cref{hoi_failures} (a), knowing that a woman’s hand is not occluding the umbrella at all may help rule out “a woman is holding an umbrella.” Additionally, in \cref{hoi_failures} (b), knowing that the skateboard in the orange box is much further than the man in the red box could help suppress “a man in the red box is riding a skateboard in the orange box.” }

\textcolor[RGB]{0,0,0}{Many other computer vision tasks, including embodied reference understanding ~\citep{chen2021yourefit} and scene de-occlusion~\citep{scenedeocclusion}, will also benefit from detecting relative distance and occlusion relationships. For example, in embodied reference understanding \cite{chen2021yourefit}, if a person is referring to an object by its relative position to another object (e.g., \emph{behind} and \emph{in front of}), being able to detect relative distance from the viewpoint can help the model understand which object that person is referring to. In scene de-occlusion, determining the relative occlusion relationship is the foundation of ordering recovery and subsequent de-occlusion of the scene~\citep{scenedeocclusion}.}

Determining relative distance relationships requires understanding the scene geometrically and deciding whether one object is closer or both are at the same distance.

\begin{figure}[t]
    \centering
    \includegraphics[width=0.8\linewidth]{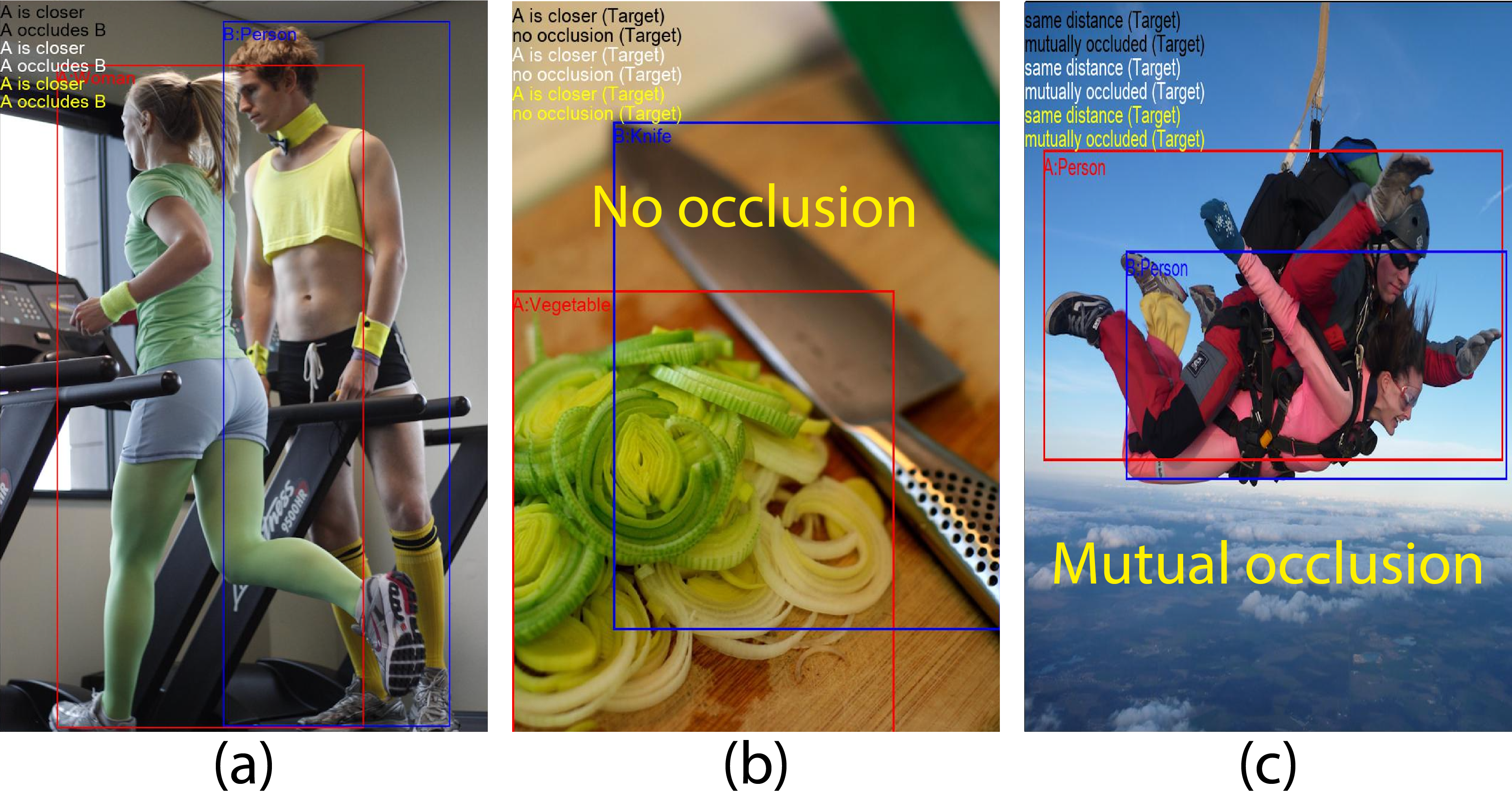}
    \caption{Examples of relative occlusion and relative distance. (a) The woman's hair and leg occlude the man, but the man does not occlude the woman. The woman is closer to the viewpoint than the man. (b) No occlusion: although the two bounding boxes intersect, a closer look at the region between vegetables and the knife would lead to the conclusion that no occlusion exists between them. (c) Mutual occlusion: the woman's arm occludes the man's arm while the man's leg occludes the woman's leg.}
    \label{teaser}
\end{figure}

Relative occlusion is more complicated. Its most straightforward form is object A occludes object B when object B does not occlude object A. For example, in \cref{teaser} (a), the woman’s hair, hip, and right leg occlude the man, but the man does not occlude the woman. Another form of relative occlusion is no occlusion. It can happen even when the bounding boxes of two objects overlap. In \cref{teaser} (b), for example, a closer look at the space between vegetables and the knife helps conclude that no occlusion exists between them. A more complicated form of occlusion is mutual occlusion. To detect mutual occlusion, a person needs to refrain from concluding one object occludes another object and keep locating additional occlusion sites. For example, in \cref{teaser} (c), the woman’s arm occludes the man’s wrist, but the relative occlusion is not the woman occludes the man. After noticing that the woman’s arm occludes the man, additional attention needs to be placed near their legs to identify that the man’s leg also occludes the woman’s leg.  

Successful detection of relative occlusion relationships between a pair of objects requires focusing attention on visual features from task-relevant regions and refraining from being distracted by irrelevant visual features. For example, when determining the relative occlusion relationship between the woman and the man in \cref{teaser} (a), the woman’s right leg and hair are more task-relevant than the window. Paying more attention to the woman’s right leg, hip, and hair is beneficial to occlusion relationship detection. In contrast, detecting occlusion relationships between these two people by attending to the window will be futile. Attention, as a selection mechanism, enables us to give higher privileges to task-relevant information and filter out distracting information~\citep{thiele2018neuromodulation}. Features from the window act as distractions and increase the complexity of identifying the relative occlusion relationship between the two people. Consequently, focused attention to task-relevant regions is critical for correctly detecting relative occlusion relationships. 

Detecting the relative distance relationship similarly requires focused attention to task-relevant regions. However, different tasks may require focused attention to different regions. For instance, detecting relative distance requires attention largely to the background to gain a geometric understanding of the scene, while detecting object pairs requires focused attention primarily to both objects to accurately locate their positions and recognize their categories. Existing artificial intelligence models, as argued by authors of~\citep{2.5vrd}, perform unsatisfactorily when detecting relative distance and relative occlusion relationships. 

Therefore, to design a model that performs well in detecting both relative occlusion and relative distance relationships, we carefully consider the importance of focused attention. Given that detecting relative occlusion and relative distance requires attention to different regions (mainly object parts v.s. objects and backgrounds), we use two separate decoders for occlusion and distance, respectively, to allow each decoder to focus on features relevant to its task without being distracted by features solely relevant to the other task. Additionally, to protect our relationship decoders from distractions from object pair detection, we use an extra decoder to propose object pairs before detecting relative relationships. 

The above considerations motivated our three-decoder model architecture in \cref{architecture}. It prevents distractions between different tasks and functions as the infrastructure for focused attention. However, the architecture alone is not sufficient for accomplishing focused attention. 

To achieve focused attention in our occlusion decoder, we use an extra task: predicting the bounding box for the region shared by (mostly when occlusions exist) or between two objects (mostly when no occlusion exists). We call this task the generalized intersection box prediction (GIT) (\cref{intersection}). 

Our GIT, a novel way to guide the model to focus on task-relevant regions, results in a more interpretable and robust system. Our experiments and attention weight visualizations demonstrate our model's ability to focus on task-specific regions when detecting relative occlusion relationships (\cref{attention_easy}) and the concomitant improvements in correctness (\cref{improvements_with_intersection} and \cref{effects_of_intersection}).

To recapitulate, 1) we propose to use the multi-decoder architecture as the infrastructure for focused attention; 2) our novel GIT  effectively guides our model to focus on occlusion-specific regions; 3) our model achieves a new state-of-the-art performance. 

\section{Problem Formulation} \label{sec:definition_of_relative_distance_and_occlusion}
\textcolor[RGB]{0,0,0}{We study the detection of relative distance and relative occlusion relationships from the viewpoint. The input is an RGB image $I$. The required outputs are the bounding box of object A (${b_A}$), the bounding box of object B ($b_B$), the relative distance relationship ($d$), and the relative occlusion relationship ($o$) between A and B.}

\subsection{Relative Distance}
\textcolor[RGB]{0,0,0}{Relative distance is from the viewpoint and is object-centric \citep{2.5vrd}. ``From the viewpoint" means the distance of an object is the length of the 3D line segment from the optical center of the input image to the object. At the same time, ``object-centric" means each object is assigned with only one distance, which is different from the ``pixel-wise depth" in monocular depth estimation where each pixel has a distance~\citep{2.5vrd}.}

\textcolor[RGB]{0,0,0}{There are four types of relative distance relationships: A is closer than B, B is closer than A, same distance, and not sure. The first two are straightforward: when the majority of human annotators in \citep{2.5vrd} believe one object is closer than the other, the relative distance is either A is closer than B or B is closer than A. }

\textcolor[RGB]{0,0,0}{For ``same distance", there is not a hard threshold. When the majority of human annotators in \citep{2.5vrd} perceive that a pair of objects are at the same distance, the relative distance relationship is ``same distance". Similarly, when the majority of the annotators are not sure about the relative distance relationship for a pair of objects, the relationship is ``not sure."}

\subsection{Relative Occlusion}
\textcolor[RGB]{0,0,0}{The relative occlusion relationship is also from the viewpoint. Occlusion ``from the viewpoint" means looking from the same direction as the camera.}

\textcolor[RGB]{0,0,0}{There are four types of relative occlusion relationships: A occludes B, B occludes A, no occlusion, and mutual occlusion. The last one, mutual occlusion, means some parts of A occlude B while some parts of B also occlude A. }

\section{Related Work}

\subsection{Models for Relative Occlusion and Distance Detection}
To the best of our knowledge, no deep learning model was designed specifically to explicitly detect relative occlusion and relative distance relationships at the same time, except multi-layer perceptrons (MLPs) in \cite{2.5vrd}. 

Models that detect relative occlusion relationships do exist \cite{scenedeocclusion}\cite{zhu2017semantic}. For example, in \cite{scenedeocclusion}, the model can determine relative occlusion relationships and build an occlusion graph, but it does not support mutual occlusion. 

Models that detect relationships between objects (visual relationship detection, or VRD) also exist \cite{ViP-CNN}\cite{PPR-FCN}\cite{VTransE}\cite{DRNet}. Unlike relative distance and relative occlusion relationship,~\citet{2.5vrd} maintain that nearly all relationships are 2D in current VRD datasets and do not address relative occlusion and relative distance relationships using the viewpoint as reference. 

To predict relative occlusion and distance relationships,~\citet{2.5vrd} designed several MLP models and modified multiple state-of-the-art VRD models to perform relative occlusion and relative distance detection. In \cite{2.5vrd}, models designed for VRD did not achieve significantly better performance than the best MLP model when detecting relative occlusion and relative distance relationships. 

In another very similar task---HOI detection---transformer models have been modified and applied by many works. Focused attention in transformer models for HOI detection has also been studied extensively in recent years. 

\subsection{Transformers for HOI detection}


HOI detection resembles relative occlusion and relative distance detection, except that relationships in HOI detection are mostly semantic. In specific, while require outputs for relative distance and occlusion relationship are ($b_a$, $b_b$, $d$, $o$) (as in \cref{sec:definition_of_relative_distance_and_occlusion}), \textcolor[RGB]{0,0,0}{those for HOI detection are ($b_h$, $b_o$, $r$, $c_o$). Here, $b_h$, $b_o$, $r$, and $c_o$ represents the bounding box of human, the bounding box of an object, the relationship between the human and the object, and the category of the object, respectively. }

Both tasks require the detection of objects and relationships. In specific, HOI detection requires the prediction of the human, the object, and the interactions between the human-object pair. Relative occlusion and distance relationship detection requires the prediction of object A, object B, and relationships between the object pair. 

The major difference lies in the relationships. In HOI, the relationships between humans and objects are mostly semantic (e.g. eat, cut, and hold). In relative occlusion and relative distance detection, however, the relationships are geometric (e.g. occludes and being closer). Additionally, semantic relationships in HOI detection are highly diverse and usually involve dozens or hundreds of relationship categories. 

Recently, many one-stage transformer-based methods have been proposed to perform HOI detection \cite{hoitr}\cite{cdn}\cite{tamura2021qpic}\cite{as-net}. These architectures first use a backbone to extract image features. The backbone is usually a convolutional neural network (CNN). Then these architectures use transformer encoders to extract global features from input images. After that, these architectures utilize transformer decoders and MLPs to produce the final predictions of bounding boxes, object classes, and relationship classes.

\subsection{Focused Attention in Transformers}
Sometimes, transformer models can naturally exhibit intense attention to task-relevant information. In object detection, decoders in DETR \cite{detr} can exhibit focused attention to the extremities of objects. In HOI detection, decoders in HOITR \cite{hoitr} can have focused attention on \textcolor[RGB]{0,0,0}{``the discriminative part"} of object pairs.

More often, however, transformer models attend heavily to a huge amount of irrelevant information. As illustrated in \cref{attention_easy}, transformers decoder allocate a significant amount of attention to many regions that are irrelevant to relative occlusion even when the only task that decoder was asked to do was detecting relative occlusion relationship. 

To let transformer models have more focused attention, many attempts have been made. \citet{correia2019adaptively} let information with nearly zero weight have exactly zero weight. \citet{zhao2019explicit} modified the self-attention module to explicitly select the elements that receive the highest attention scores and explicitly let their transformer model focus on these elements. \citet{zhu2020deformable} designed their \textcolor[RGB]{0,0,0}{``deformable attention module,"} allowing their transformer model to have focused attention on a small number of pixels near a reference point.

It can be insufficient, however, to just focus on elements with the highest attention scores. In some complex tasks composed of multiple sub-tasks, relevant information to one sub-task can be distracting to another sub-task. Task-specific attention, in this case, is difficult to achieve due to the distinctions between different sub-tasks \cite{cdn}. For example, in HOI detection, many locations within or near a human-object pair can be highly relevant for detecting the human and the object, but regions relevant to interaction detection can vary a lot based on the category of a specific interaction \cite{hoitr}. In other words, some information that is relevant to human-object pair detection can at the same time be distracting to interaction detection and vice versa.

To allow focused attention for each sub-task, researchers used separate decoders for different sub-tasks and connect them in parallel \cite{as-net} or serial\cite{cdn} manners. In specific, \citet{as-net} proposed to use a parallel connection of two separate transformer decoders for two sub-tasks, and introduced an attention module between each layer of the two separate decoders to aggregate relevant information from one to the other. Additionally, \citet{zhang2021mining} proposed to use a series connection of separate transformer decoders for different sub-tasks, allowing each decoder to focus on only one task and attending to information that is relevant to that one task. 

In the following section, we will introduce our method for achieving focused attention in the occlusion decoder. Specifically, we use a combination of series and parallel connections of transformer decoders to avoid distractions from different sub-tasks and use a generalized intersection box prediction task to further guide our occlusion decoders to attend to task-relevant regions.

\section{Method}

\begin{figure*}[t]
  \centering
  \includegraphics[width=0.8\linewidth]{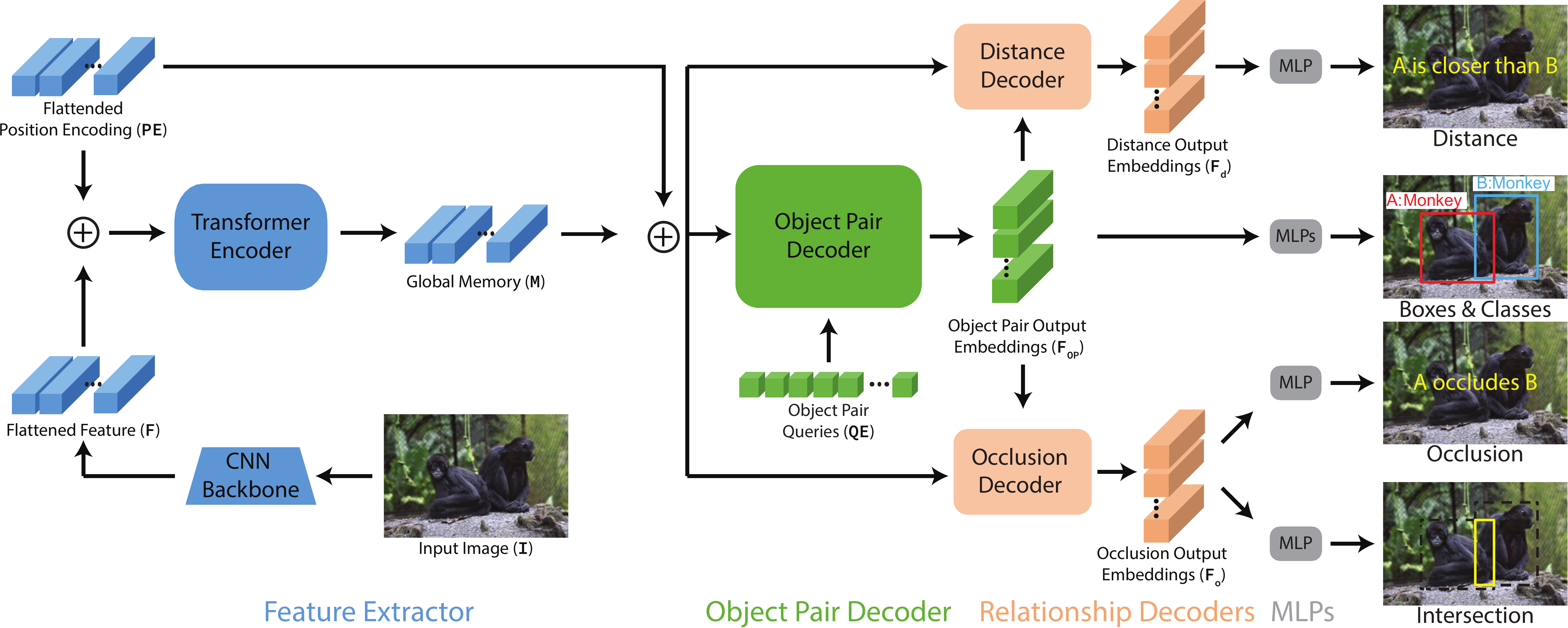}
  \caption{Model Architecture. Our model is comprised of a feature extractor, an object pair decoder, two relationship decoders, and multiple MLP heads. The features extractor uses a CNN backbone and a transformer encoder to extract global visual features \textbf{M}. Using \textbf{M}, the object pair decoder produces object pair embeddings $\mathbf{F}_{OP}$. Using $\mathbf{F}_{OP}$ as queries, two relationship decoders parse relative occlusion and relative distance relationships, respectively. The outputs of relationship decoders are $\mathbf{F}_{d}$ and $\mathbf{F}_{o}$. Finally, taking the output embeddings of three decoders, MLP heads produce the final predictions of object boxes, object classes, relative distances, relative occlusions, and generalized intersection boxes.}
  \label{architecture} 
\end{figure*}

In this section, we describe the model architecture, matching strategy, and loss computation. Our model architecture has four components: feature extractor, object pair decoder, relationship decoders, MLP heads for classification and regression. 

\subsection{Overview}
Our feature extractor utilizes a CNN backbone and a transformer encoder to extract context-aware visual features from input RGB images. Extracted features are fed into our object pair decoder for bounding box and class label predictions. Taking the outputs of our object pair decoder as queries, our occlusion and distance transformer decoders predict relative occlusion and relative distance relationships, respectively. Our occlusion decoder also predicts generalized intersection boxes. An illustration of our proposed model architecture is in \cref{architecture}.

We use transformer because it models \textcolor[RGB]{0,0,0}{long-range dependencies}, which is critical for our task. \textcolor[RGB]{0,0,0}{Specifically, the transformer has a global receptive field, which helps detect the relative distance and occlusion relationships when two objects are very far away from each other. Additionally, when two objects are not very far from each other, the transformer's global receptive field will also help detect relative distance relationships because it allows a more comprehensive consideration of the spatial configurations of the scenes. CNNs, in contrast, according to \citet{luo2016understanding}, have a relatively small effective receptive field in which an input pixel can have a ``non-negligible impact" on the output. Therefore, we choose the transformer model instead of CNNs for its greater capacity to capture information from a larger area. }

\subsection{Feature Extractor}
Our visual feature extractor consists of a CNN backbone and a transformer encoder. It models long-range dependencies and extracts context-aware visual features for downstream tasks. 

\subsubsection{CNN Backbone} 
Given an RGB image $\mathbf{I}$ of shape $(3, H, W)$, a CNN backbone (e.g., ResNet-101 \cite{resnet}) produces a feature map $\textbf{F}_0$ of shape $(C, h, w)$, where $C$ is the number of output channels of CNN backbone, $h=\lceil{\frac{H}{32}}\rceil$, and $w=\lceil{\frac{W}{32}}\rceil$, where $\lceil\cdot\rceil$ is the ceiling function. We reduce the dimension of the feature map from $C$ to $d$ using a $1 \times 1$ convolutional layer with stride $1$, producing reduced feature of shape $(d, h, w)$. We then flatten the reduced feature and generate flattened feature $\mathbf{F}$ of shape $(hw,d)$.\\

\subsubsection{Transformer Encoder}
Our transformer encoder takes the flattened feature $\mathbf{F}$ as input. It is composed of $N_l$ transformer encoder blocks. Each block is composed of a multi-head self-attention layer and a feed-forward layer. Inside each self-attention layer, there are $N_h$ heads. The output  $A$ of each self-attention layer is formed by the concatenation of outputs $A_i$'s from all $N_l$ heads. 

Before feeding the flattened feature $\mathbf{F}$ into our transformer encoder, however, we generate position encoding \cite{positionencoding1}\cite{positionencoding2} $\mathbf{PE}$ to facilitate the use of relative position information in downstream modules. The shape of $\mathbf{PE}$ is $(hw,d)$, where $d$ is the hidden dimension of our transformer encoder. 

After obtaining $\mathbf{PE}$ is $(hw,d)$, we feed $\mathbf{F}$ and $\mathbf{PE}$ into the first transformer encoder block. 

In each block, we denote the input feature as $\mathbf{F}_{in}$. The query $\mathbf{Q}$, key $\mathbf{K}$, and value $\mathbf{V}$ of each transformer encoder block are 
\begin{align}
\begin{split}
    (\mathbf{Q}, \mathbf{K},\mathbf{V}) &= (\mathbf{F}_{in}+\mathbf{PE},\ \mathbf{F}_{in}+\mathbf{PE}, \ \mathbf{F}_{in})\\ 
\end{split}    
\end{align}

For each head $i=1,2, \cdots, N_h$ inside a self-attention layer, the attention weights $W^Q_i$, $W^K_i$ and $W^V_i$ all have shape $(d, \frac{d}{N_{h}})$. For each head, the query $\mathbf{Q}_i$, key $\mathbf{K}_i$, and value $\mathbf{V}_i$ are
\begin{align}
\begin{split}
    (\mathbf{Q}_i, \mathbf{K}_i, \mathbf{V}_i)  &= (\mathbf{Q} W^Q_{i}, \mathbf{K} W^K_{i}, \mathbf{V} W^V_{i})
\end{split}    
\end{align}
Each head produces an output $A_i$
\begin{align}
\begin{split}
    A_{i} &= \Big[\text{Softmax}\Big  (\frac{\mathbf{Q}_i\mathbf{K}_i^T}{\sqrt{d}}\Big)\mathbf{V}_i\Big]\\
\end{split}    
\end{align}
The outputs of all heads in a self-attention layer are concatenated to form $A$
\begin{align}
\begin{split}
    A &= \text{Concat}(A_1, A_2, \cdots, A_{N_h})\\
\end{split}    
\end{align}
After obtaining the concatenated $A$, the final output $\mathbf{F}_{attn}$ of a self-attention layer is generated using a linear layer with dropout and a residual connection with input feature $\mathbf{F}_{in}$
\begin{align}
\begin{split}
    \mathbf{F}_{attn} &= \text{Linear}(\text{Dropout}(A)+\mathbf{F}_{in}) \in \mathbb{R}^{hw\times d}
\end{split}    
\end{align}

After obtaining $\mathbf{F}_{attn}$ from the self-attention layer inside each transformer encoder block, $\mathbf{F}_{attn}$ is fed into the feed-forward layer (FFN), which is composed of two fully-connected linear layers with dropout and ReLU activation. FFN is applied to each feature vector in $hw$ positions identically and individually and outputs a total feature $\mathbf{F}_{out}$ of the same shape $(hw,d)$, which will serve as input to the self-attention layer in the next next block. 

Finally, the output $\mathbf{F}_{out}$ of the last transformer encoder block is denoted as global memory $\mathbf{M}$ of shape $(hw,d)$.

\subsection{Object Pair Decoder}
Our object pair decoder detects object pairs. It takes $\mathbf{M}$, $\mathbf{PE}$, and object pair queries $\mathbf{QE}$ as inputs. It is composed of $N_{pair}$ transformer decoder blocks. Each transformer decoder block is composed of a multi-head self-attention module and a multi-head cross-attention module followed by a feed-forward network. The attention mechanism is the same as that in encoder layers, so we only specify our choice of $\mathbf{Q},\mathbf{K}$ and $\mathbf{V}$ in the self-attention and cross-attention layers. 

To enable proposing a fixed number of object pairs, we choose a hyper-parameter $N_q$ as the number of object pairs queries to be outputted by the decoder, and we generate parameterized queries with $N_q$ learnable embeddings.

\subsubsection{Self-attention Module}
For $k=1,2 ... N_{pair}$, we denote the input of the $k$-th self-attention module as $\mathbf{F}^k_{in}$ and the output of the $k$-th FFN as $\mathbf{F}^k_{out}$. Both of them have shape $(N_q,d)$.
\begin{align}
\mathbf{F}^k_{in} = \mathbf{F}^{k-1}_{out} + \mathbf{QE} 
\end{align}
where QE is a set of learnable parameters of shape ($N_{pair}$,d). We set $\mathbf{F}^{0}_{out} = \mathbf{0}$ for consistency of notation below. In the $k$-th self-attention module ($k=1,\cdots,N_{pair}$), the query $\mathbf{Q}$, key $\mathbf{K}$, and value $\mathbf{V}$ are 
\begin{align}
(\mathbf{Q}, \mathbf{K}, \mathbf{V}) = (\mathbf{F}^k_{in}, \mathbf{F}^k_{in}, \mathbf{F}^{k-1}_{out})
\end{align}

\subsubsection{Cross-attention Module} The $k$-th self-attention module outputs a feature $\mathbf{F}^k_{attn}$ with shape $(N_q,d)$ to be fed into the $k$-th cross-attention module. Together with the global memory $\mathbf{M}$ from the transformer encoder and the fixed positional encoding $\mathbf{PE}$, we let
\begin{align}
(\mathbf{Q}, \mathbf{K}, \mathbf{V}) = (\mathbf{F}^k_{attn},\ \mathbf{M}+\mathbf{PE},\ \mathbf{M})
\end{align}

On top of the cross-attention layer, the FFN layer is the same as that in the encoder architecture. The $k$-th FNN layer outputs $\mathbf{F}^k_{out}$, which is also the overall output embedding of the $k$-th decoder block. Finally, we stack $\mathbf{F}^k_{out}$ to obtain object pair output embeddings $\mathbf{F}_{op}$ of shape $(N_{pair}, N_{q}, d)$
\begin{align}
\mathbf{F}_{op} = \text{Stack}(\mathbf{F}^k_{out},\text{ }k=1,\cdots, N_{pair})    
\end{align}

\subsection{Relationship Decoders}
The relationship decoders include the relative occlusion decoder and relative distance decoder. Both decoders share the same structure with the object pair decoder. That is, each decoder block contains self-attention, cross-attention, and feed-forward modules. For the self-attention module and the cross-attention module, as downstream decoders, they use $\mathbf{F}^{N_{pair}}_{out}$, the output of the last object pair decoder layer as queries. In particular, we choose
\begin{align}
    (\mathbf{Q}, \mathbf{K}, \mathbf{V}) = (\mathbf{F}^{N_{pair}}_{out},\ \mathbf{M}+\mathbf{PE},\ \mathbf{M})
\end{align}
The outputs of the relationship decoders are distance output embeddings $\mathbf{F}_{d}$ and occlusion output embeddings $\mathbf{F}_o$. Both of them have shape $(N_{q},d)$. 

\subsection{MLP Heads}
The output embeddings $\mathbf{F}_{op}$, $\mathbf{F}_{d}$, and $\mathbf{F}_{o}$ are fed into separate classification MLP heads and box regression MLP heads to produce the final predictions of instances in the input image. All MLP heads have the same structure except the number of nodes in the last layer, where each classification MLP head produces a probability vector via softmax over each class, while each regression MLP head produces a four-vector $(c_x,c_y,w,h)\in [0,1]^4$ via sigmoid function. Here $(c_x, c_y)$ is the relative coordinate of the center of box and $(w,h)$ denotes its width and height (see \cref{MLP}). 

\textcolor[RGB]{0,0,0}{Our model treats different types of relative distance relationships as different classes so there is not an explicit threshold between ``same distance" and one object being closer than the other. In other words, our model predicts two objects are at the ``same distance" when the logit for the ``same distance" class is the greatest among those for all four. }

\begin{table}[t]
\caption{Prediction MLP Heads: Notations (NOT) and Shapes}
\centering
\label{MLP}
\resizebox{0.65\linewidth}{!}{
\begin{tabular}{llll}
\toprule
 \textbf{MLP \#} & \textbf{Prediction} & \textbf{NOT}  & \textbf{Shape}\\ 
\midrule
Object Pair MLP-1 & Object A Class          & $p_A$     & $(N_{pair}, N_c)$ \\
Object Pair MLP-2 & Object B Class & $p_B$     & $(N_{pair}, N_c)$ \\ 
Object Pair MLP-3 & Object A Box            & $p_{b_A}$ & $(N_{pair}, 4)$   \\ 
Object pair MLP-4 & Object B Box            & $p_{b_B}$ & $(N_{pair}, 4)$   \\ 
Distance MLP     & Dist. Predicates      & $p_d$     & $(N_{pair}, N_d)$ \\ 
Occlusion MLP-1   & Occl. Predicates     & $p_o$     & $(N_{pair}, N_o)$ \\ 
Occlusion MLP-2   & Intersection Box        & $p_{int}$ & $(N_{pair}, 4)$   \\
\bottomrule
\label{MLP_table}
\end{tabular}
}
\end{table}

\subsection{Generalized Intersection Box Prediction Task}
With GIT, our occlusion decoder produces an additional generalized intersection box for each pair of objects. First \textcolor[RGB]{0,0,0}{we denote the bounding box of objects $A$ and $B$ as}
\begin{align}
\begin{split}
    b_A &= [x_A, X_A] \times [y_A, Y_A]\\
    b_B &= [x_B, X_B] \times [y_B, Y_B]
\end{split}
\end{align}
where $\times$ denotes the Cartesian product, $(x_i,y_i)$ is the coordinates of bottom left corner of $b_i$, and $(X_i,Y_i)$ is the top right corner of $b_i$, for $i\in\{A, B\}$. Then the generalized intersection box $b_{\cap}$ for a pair of objects A and B is defined as
\begin{align}
    b_{\cap} = [x_\cap, X_\cap] \times [y_\cap, Y_\cap]
\end{align}
where $x_\cap$ and $X_\cap$ are the second and third smallest elements in $\{x_A, X_A, x_B, X_B\}$ and $y_\cap$ and $Y_\cap$ are the second and third smallest elements in $\{y_A, Y_A, y_B, Y_B\}$.

\begin{figure}[t]
    \centering
    \includegraphics[width=1\linewidth]{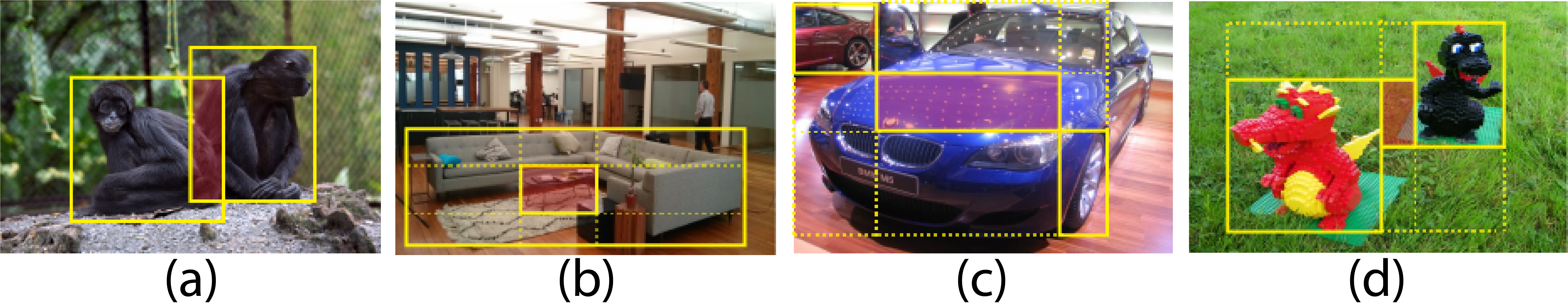}
    \caption{Generalized Intersection Box (shown in red). (a) When two bounding boxes partially intersect, the generalized intersection box is the same as the intersection box. (b) When a smaller bounding box is completely inside a larger bounding box, the generalized intersection box is the same as the smaller bounding box. (c) and (d) When two bounding boxes do not intersect, the generalized intersection box is formed using the second and third smallest x coordinates and y coordinates of the two bounding boxes.}
    \label{intersection}
\end{figure}

As shown in \cref{intersection} (a), the generalized intersection box is the overlapping region of the bounding boxes of two apes, and in (b), the bounding box of the tea table is contained in the bounding box of the sofa, so the overlapping region is the bounding box of the tea table itself. In (c) and (d), there is no intersection between object bounding boxes: the red car at the upper left corner and the wheel of the blue car at the lower right corner in (c), and two toys on the grassland in (d). In these cases, we take the middle box that shares sides (or extended sides) with the bounding boxes of $A$ and $B$. From a general perspective, the generalized intersection box is an important link between $A$ and $B$, regardless of whether the bounding boxes of $A$ and $B$ actually intersect.

\subsection{Matching Strategy}
Since each image contains different numbers of ground truth instances, which is usually smaller than the number of predictions generated by the model $N_q$ (equals 100 in our implementation), we need to find a strategy to assign each ground truth to a prediction. Since each of the training images contains only labels for one pair of objects, whereas many other instances exist but are unlabelled, most of the predictions concerning the unlabelled instances should not be trained. Hence it is important that we only match the best prediction with the ground truth labels while viewing other predictions as background. We adapted the matching strategy in \cite{hoitr} as follows. For each training image, we assign the ground-truth targets, denoted as $\{g^1, g^2, \cdots, g^{N_t}\}$, where $N_{t}$ is the number of targets, to predictions $\{p^1, p^2, \cdots, p^{N_q}\}$ using bipartite matching. That is, we seek for a one-to-one function $\sigma:\{1,2,\cdots, N_t\} \to \{1,2,\cdots,N_q\}$, so that $g^i$ is matched to $p^{\sigma(i)}$. Then we will minimize the cost among all possible matching functions $\sigma\in\mathcal{S}$
\begin{align}
\sigma_0 = \argmin_{\sigma\in\mathcal{S}}\sum_{i=1}^{N_t} C(g^i, p^{\sigma(i)})    
\end{align}
In order to get the most efficient training, we assign targets in the way that minimizes the matching cost $C$ using the Hungarian algorithm \cite{kuhn1955hungarian}. The computation of the matching cost $C(g,p)$ for a certain matched pair of target $g$ and prediction $p$ is as follows. $C(g,p)$ is a weighted sum of classification cost $C^{c}(g_c,p_c)$ and regression cost $C^{r}(g_r,p_r)$:
\begin{align}
    C(g,p) = \beta_{c} C^{c}(g, p) + \beta_{r} C^{r}(g, p)
    \label{cost_total}
\end{align}
Since the classification outputs are the confidence of each potential class (of objects and relationship predicates), $C^{c}$ measures the closeness of the model's confidence prediction of the ground-truth class to $1$, that is:
\begin{align}C^c(g_i,p_i) = 1 - p_i[k]
\end{align}
where $k$ is the ground-truth class.

Using hyper-parameters $\alpha_{A}$, $\alpha_{B}$, $\alpha_{d}$ and $\alpha_{o}$, where $d$ and $o$ refer to distance and occlusion classes respectively. We compute the total classification cost as a normalized weighted sum of $C^c(g_i,p_i)$ for $i\in \{A,B,d,o\}$:
\begin{align}
    C^{c}(g,p) &=\Big[\sum_{i} \alpha_i C^{c}(p_i,g_i)\Big]\Big/\Big[\sum_{i} \alpha_i\Big]
    \label{cost_class}
\end{align}

The regression cost of each predicted box is a weighted sum of $l_1$ cost and generalized IoU (GIoU) cost. For each $j\in\{b_A,b_B,int\}$, where $int$ is the generalized intersection box, we let
\begin{align}
\label{Regression Cost}
    C^{r}(g_j,p_j) = \alpha_{l_1}\cdot\|g_j- p_j\|_{1} + \alpha_{giou}\cdot\text{GIoU}(g_{j}, p_j)
\end{align}
and if model predicts generalized intersection box, let
\begin{align}
C^r(g,p) = \frac13\sum_{j}C^r(g_j,p_j),\quad j\in\{b_A, b_B, int\}    
\end{align}
or if the generalized intersection box is not predicted, let
\begin{align}
C^r(g,p) = \frac{1}{2}\sum_{j}C^r(g_j,p_j),\quad j\in\{b_A, b_B\}
\end{align}

\textcolor[RGB]{0,0,0}{The generalized intersection box is always predicted when using GIT and always not predicted when not using GIT. One exception is the ablation study in \cref{sec:experiment_predicte_intersection_only_when_intersection_exists}, in which our model only predicts the generalized intersection box when two bounding boxes intersect with each other. }





\subsection{Loss Computation}
The training loss is based on the result of matching, and its computation is slightly different from the cost function between matched ground truth-prediction pairs. Apart from the given classes of foreground objects in an image, we consider all objects not in the ground truth targets as a single class $O_{bg}$ called background object. The background object is encoded as a one-hot vector $\mathbf{e}_{N_c+1}$ without a bounding box, where $N_{c}$ is the number of instance classes.

\subsubsection{Classification loss} The classification loss for objects and relationships are Negative Log Likelihood (NLL) loss. For $i\in\{A,B,d,o\}$ and matched pairs $(g,p)$, we use weight parameters $\alpha_i$:
\begin{align}
L^c(g_i, p_i) = -\alpha_i\cdot \log(p_i[k])
\end{align}
where $k$ is the ground truth class. For $i\in\{A,B\}$ and unmatched prediction $p$, we use a small weight parameter $\alpha_{eos}$ to control the background loss
\begin{align}
L^c(O_{bg}, p_i) = -\alpha_{eos}\cdot\log(p_i[N_c + 1])    
\end{align}
\subsubsection{Regression loss} The regression loss are only applied to matched pairs $(g,p)$, because background object does not have a bounding box to be compared with. For $j\in\{A,B,int\}$, we use the same formula as \cref{Regression Cost}:
\begin{align}
L^{r}(g_j,p_j) = \alpha_{l_1}\cdot\|g_j- p_j\|_{1} + \alpha_{giou}\cdot\text{GIoU}(g_{j}, p_j)    
\end{align}
\subsubsection{Total loss} Finally, the total loss is again weighted by $\beta_c$ and $\beta_r$:
\begin{align}
\begin{split}
L_{total}= &\beta_c\bigg[ \sum_{(g,p)}^\text{matched} L^c(g,p) +\sum_{p}^\text{unmatched} L^c(O_{bg},p)\bigg]\\
&+ \beta_r \sum_{(g,p)}^\text{matched} L^r(g,p)
\end{split}
\end{align}

\subsection{Inference}
To make a final inference of instances based on the model output, we first perform the non-maximum suppression (NMS) to filter out duplicate predictions. In particular, we first sort the list of pairs in decreasing order in overall confidence measured by the product of classification confidence of distance and occlusion:
\begin{align}
\label{Confidence}
    \text{Conf}(p) = \|p_d\|_\infty \cdot \|p_o\|_\infty
\end{align}
For two predictions $p^1$ and $p^2$, we regard them as duplicate predictions if they satisfy the following similarity criterion:
\begin{align}
\text{IoU}(p^1_{b_A}, p^2_{b_A}) > 0.7 \quad\text{and}\quad \text{IoU}(p^1_{b_B}, p^2_{b_B}) > 0.7  
\end{align}
and if the predicted object categories and relative relationships in $p^1$ and $p^2$ match. We will only keep the prediction with the highest overall confidence Conf(p) when there are duplications.

\textcolor[RGB]{0,0,0}{This duplication removal strategy does not consider two predictions, ``A is closer than B" and ``B is further than A," as duplications because their relationships are different (closer v.s. further). Additionally, unless the bounding boxes of both objects significantly overlap, object A in the former prediction will not have greater than 0.7 IoU with object A in the latter one, not meeting the IoU threshed for being considered as a pair of duplications. Furthermore, unless both objects have the same predicted category, they will not meet the object category requirement for being duplications. Therefore, most of the time, our duplication removal strategy will not consider predictions such as ``A is closer than B" and ``B is further than A" for the same pair of objects as duplications. Our model is required to predict both of them because our ground truth contains both. }

After filtering out duplicate predictions, we get a list of predictions as our result of inference: $\{p^1,p^2,\cdots,p^N\}$.

\section{Experiments}  
We conduct experiments to evaluate the performance of our model in detecting relative occlusion and relative distance relationships. We report F1-score (F1), precision (p), and recall (r); and discuss model performance boost by referring to F1-scores. To demonstrate the performance of our method, we also provide extensive qualitative results on images from the 2.5VRD validation and test set and video frames from in-the-wild videos. We also study the effects of GIT on occlusion decoder attentions through ablation experiments and provide qualitative comparisons. \textcolor[RGB]{0,0,0}{Additionally, we investigate the benefits of predicting the generalized intersection when no intersection exists and using separate decoders for different sub-tasks. }

\subsection{Dataset}
The 2.5VRD dataset \cite{2.5vrd} contains annotated images for the within-image relative occlusion and relative distance detection task and image pairs for the across-image relative distance detection task. We conduct experiments using the within-image portion of the dataset. It contains 105660, 1196, and 3987 images in the training, validation, and test set, respectively. Annotations for the training set are very sparse. Only one pair of objects is annotated for each training image. Annotations for the validation and test sets, in contrast, are exhaustive.

\subsection{Evaluation Metric}
During the evaluation, we leak the number of annotated objects to our model and require our model to accurately detect object pairs and correctly predict corresponding relationships.

\subsubsection{Leaking the Number of Objects}
Using the same evaluation metric used in \cite{2.5vrd}, we leak n, the number of annotated objects in an image, to our model. Knowing how many objects are in the annotations, our model then outputs at most n(n-1) predictions for evaluation. However, the maximum number of predictions our model can produce is limited by $N_q$, and duplicated predictions are filtered out using non-maximum suppression. As a result, knowing the number of objects in annotations and limited by the number of predictions it can produce, our model would finally produce $\min\{n(n - 1), N\}$ predictions for evaluation, where N is the number of remaining predicted pairs after non-maximum suppression. 

\begin{table}[t]
    \caption{Comparison to state-of-the-art methods. Our model outperforms the previously best model by 4.8\% and 7.0\% on relative distance and relative occlusion, respectively.}
    \centering
    \resizebox{0.8\linewidth}{!}{
    \begin{tabular}{ lllll } 
        \hline
        & Dist F1 & Occl F1 & Dist p / r & Occl p / r\\ 
        \hline
        ViP-CNN \cite{ViP-CNN} & 33.6 & 34.2 & 33.3 / 33.9 & 34.1 / 34.3\\
        PPR-FCN \cite{PPR-FCN} & 33.5 & 33.9 & 33.0 / 34.1 & 33.8 / 34.0\\
        VTransE \cite{VTransE} & 32.4 & 32.9 & 31.5 / 33.2 & 32.8 / 33.0\\
        DRNet \cite{DRNet} & 33.8 & 34.4 & 33.9 / 33.7 & 34.3 / 34.5\\
        \textbf{Ours} & \textbf{38.6} \textcolor[RGB]{205,0,0}{(+14\%)} & \textbf{41.2} \textcolor[RGB]{205,0,0}{(+20\%)} & \textbf{38.9} / \textbf{38.2} &\textbf{41.6} / \textbf{40.9}\\
        \hline
    \end{tabular}
    }%
    \label{comparison_to_state_of_the_art}
\end{table}

\begin{table}[t]
    \caption{Effects of varying the number of object pair decoder layers.}
    \centering
    \resizebox{0.7\linewidth}{!}{
    \begin{tabular}{ clllll } 
        \toprule 
        $N_{pair}$ & Dist F1 & Occl F1 & Dist p / r & Occl p / r\\ 
        \midrule
        3 & 31.3 \textcolor[RGB]{0,200,0}{(-19\%)} & 33.9 \textcolor[RGB]{0,200,0}{(-18\%)} & 31.7 / 30.9 & 34.3 / 33.5\\
        6 & \textbf{38.6} & \textbf{41.2} & \textbf{38.9} / \textbf{38.2} & \textbf{41.6} / \textbf{40.9}\\
        9 & 26.7 \textcolor[RGB]{0,200,0}{(-31\%)} & 28.9 \textcolor[RGB]{0,200,0}{(-30\%)} & 27.0 / 26.5 & 29.1 / 28.6\\
        \bottomrule
    \end{tabular}
    }
    \label{varying_num_decoder_layers_in_object_pair_decoder}
\end{table}

\begin{table}[t]
    \caption{Effects of varying the number of distance decoder layers.}
    \centering
    \resizebox{0.7\linewidth}{!}{
    \begin{tabular}{ cllll } 
        \toprule
        $N_{d}$ & Dist F1 & Occl F1 & Dist p / r & Occl p / r\\ 
        \midrule
        1 & 35.2 \textcolor[RGB]{0,200,0}{(-9\%)} & 37.7 \textcolor[RGB]{0,200,0}{(-8\%)} & 35.6 / 34.9 & 38.2 / 37.3\\
        3 & \textbf{38.6} & \textbf{41.2} & \textbf{38.9} / \textbf{38.2} & \textbf{41.6} / \textbf{40.9}\\
        6 & 37.3 \textcolor[RGB]{0,200,0}{(-3\%)} & 40.0 \textcolor[RGB]{0,200,0}{(-3\%)} & 37.6 / 36.9 & 40.4 / 39.6\\
        \bottomrule
    \end{tabular}
    }
    \label{varying_num_decoder_layers_in_distance_decoder}
\end{table}

\begin{table}[t]
    \caption{Effects of varying the number of occlusion decoder layers.}
    \centering
    \resizebox{0.7\linewidth}{!}{
    \begin{tabular}{ cllll } 
        \toprule
        $N_{o}$ & Dist F1 & Occl F1 & Dist p / r & Occl p / r\\ 
        \midrule
        1 & 36.6 \textcolor[RGB]{0,200,0}{(-5\%)} & 38.6 \textcolor[RGB]{0,200,0}{(-6\%)} & 37.0 / 36.2 & 39.0 / 38.2\\
        3 & \textbf{38.6} & \textbf{41.2} & \textbf{38.9} / \textbf{38.2} & \textbf{41.6} / \textbf{40.9}\\
        6 & 37.7 \textcolor[RGB]{0,200,0}{(-2\%)} & 40.4 \textcolor[RGB]{0,200,0}{(-2\%)} & 38.0 / 37.4 & 40.7 / 40.0\\
        \bottomrule
    \end{tabular}
    }
    \label{varying_num_decoder_layers_in_occlusion_decoder}
\end{table}

\subsubsection{Correct Detection}
As in \cite{2.5vrd}, a prediction $p$ is considered as a correct detection if both bounding boxes $p_{b_A}, p_{b_B}$ have greater than 0.5 IoU with respect to ground truth bounding boxes. 

\subsubsection{Correct Prediction}
A correct detection, if its predicates $p_d,p_o$ for both relative distance and occlusion are correct, it is considered a correct prediction. Note that, as in \cite{2.5vrd}, we do not take predicted object categories into consideration. To discourage duplicate detection, when counting true positive (TP) predictions, if a ground-truth instance is matched to multiple predictions, then all these predictions except the first one being matched are considered false positives (FP). Hence a prediction is considered as TP for a ground truth if and only if it is a correct prediction and it is the first one being matched with the corresponding ground truth.

\subsection{Implementation Details}
We apply data augmentation to images. For images in the training set, we apply random horizontal flip, random resize, and random adjustments of brightness and contrast. After applying data augmentation, we normalize all images. 

We use the Res101 backbone in \cite{hoitr} pre-trained on a different dataset than 2.5VRD. The number of transformer encoder layers is 6. The number of queries in the object pair decoder is 100. We set the learning rate to 1e-4 and drop it once to 1e-5 at the 30th epoch. We use the Adam optimizer and train all models for 40 epochs. The dropout rate is 0.1. The weight for losses of background pairs is 0.02. 

\textcolor[RGB]{0,0,0}{In \cref{cost_total}, classification cost weight $\beta_c$ and regression cost $\beta{r}$ is 1.2 and 1.0, respectively. In \cref{cost_class} class cost weight for object A ($\alpha_A$), object B ($\alpha_B$), distance ($\alpha_d$) , and occlusion ($\alpha_o$) is 1, 1, 2, and 2, respectively. In \cref{Regression Cost}, the regression cost weight for $l_1$ cost ($\alpha_{l_1}$) and GIoU cost is ($\alpha_{giou}$) is 5 and 2, respectively.}

During training, we sort predictions by the product of relationship confidences. When generating attention weight visualizations (\cref{attention_easy} and \cref{improvements_with_intersection}), however, we sort predictions by the product of relationship confidences and object class confidences.

Photos and visualizations used in this paper may contain object pairs that are predicted by our models but are not in the ground truth annotations. Nonetheless, the occlusion relationships between these object pairs that are not in the ground truth are nonambiguous.

\subsection{Comparisons with State-of-the-Art Methods}
We compare the performance of our model with state-of-the-art VRD models\cite{ViP-CNN}\cite{PPR-FCN}\cite{VTransE}\cite{DRNet} evaluated by \cite{2.5vrd}. On the 2.5VRD dataset, our model achieves 38.6\% and 41.2\% F1-score on relative distance and relative occlusion, respectively (\cref{comparison_to_state_of_the_art}). Our model outperforms the previously best model--DRNet--14\% and 20\% on relative distance and relative occlusion, respectively.

\subsection{Ablation Studies}

We conduct ablation studies to investigate the effects of decoder layer numbers, the generalized intersection box prediction task, predicting the generalized intersection box when no intersection exists, and using separate decoders for different sub-tasks on model performance. 

\subsubsection{Number of Decoder Layers}
We study the effects of varying the number of transformer decoder layers in our object pair decoder, distance decoder, and occlusion decoder. Unless otherwise specified, we use 6, 3, and 3 decoder layers for object pair, distance, and occlusion, respectively. 

We observe significant decreases in model performance when varying the number of object pair decoder layers (\cref{varying_num_decoder_layers_in_object_pair_decoder}). Specifically, decreasing the number of layers from 6 to 3 results in a 19\% drop in distance F1-score and an 18\% drop in occlusion F1-score. Increasing the number of layers from 6 to 9 leads to a 31\% drop in distance F1-score and a 30\% drop in occlusion F1-score. Using the object pair decoder, we aim to extract object pair features that serve as queries for downstream tasks. The significant performance decrease verifies the importance of these queries and indicates that the object pair decoder may also involve feature extractions for depth and occlusion. These results show that a 3-layer decoder cannot provide sufficient feature extraction for downstream tasks, and a 9-layer decoder may lead to an overfitting result.

In the distance decoder and occlusion decoder, varying the number of transformer layers results in minor drops in model performance (\cref{varying_num_decoder_layers_in_distance_decoder} and \cref{varying_num_decoder_layers_in_occlusion_decoder}). Specifically, decreasing the number of distance layers from 3 to 1 results in a 9\% drop in distance F1-score and an 8\% drop in occlusion F1-score. Increasing the number of distance layers from 3 to 6 results in a 3\% drop in distance F1-score and a 3\% drop in occlusion F1-score. Similarly, decreasing the number of occlusion layers from 3 to 1 causes a 5\% drop in distance F1-score and a 6\% drop in occlusion F1-score. Increasing the number of occlusion layers to 6 causes a 2\% drop in distance F1-score and a 2\% drop in occlusion F1-score. Likewise, simultaneously varying the number of distance decoder layers $N_{d}$ and occlusion decoder layers $N_{o}$ (\cref{varying_num_decoder_layers_in_both_decoders}) results in minor drops in model performance. These results show a 3-layer decoder can provide sufficiently fine features for classification and regression, and increasing or decreasing the number of layers leads to worse feature extraction results. At the same time, these smaller performance drops, compared to the significant ones when changing the number of object pair decoder layers, further indicates the more vital role the object pair decoder plays for downstream tasks.


\begin{table}[t]
    \caption{Effects of simultaneously varying the number of distance and occlusion decoder layers.}
    \centering
    \resizebox{0.7\linewidth}{!}{
    \begin{tabular}{ ccllll } 
        \toprule
        $N_{d}$ & $N_{o}$ & Dist F1 & Occl F1 & Dist p / r & Occln p / r\\ 
        \midrule
        1 & 1 & 36.5 \textcolor[RGB]{0,200,0}{(-5\%)} & 39.0 \textcolor[RGB]{0,200,0}{(-5\%)} & 36.9 / 36.1 & 39.4 / 38.6 \\
        3 & 3 & \textbf{38.6} & \textbf{41.2} & \textbf{38.9} / \textbf{38.2} & \textbf{41.6} / \textbf{40.9}\\
        6 & 6 & 38.1 \textcolor[RGB]{0,200,0}{(-1\%)} & 40.8 \textcolor[RGB]{0,200,0}{(-1\%)} & 38.4 / 37.8 & 41.2 / 40.5\\
        \bottomrule
    \end{tabular}
    }
    \label{varying_num_decoder_layers_in_both_decoders}
\end{table}

\subsubsection{Generalized Intersection Box Prediction Task}
We study the effects of the generalized intersection box prediction task on model performance (\cref{effects_of_intersection}). Since changing the number of object pair decoder layers would result in significant performance drops, we fix the number of object pair decoder layers to 6 and vary the number of decoder layers in distance and occlusion decoders to form three different settings. Under all these three settings, our proposed GIT leads to performance boosts on both distance and occlusion F1-scores. 

\begin{table}[t]
    \caption{Effects of the generalized intersection box prediction task (GIT). (Inter: intersection)}
    \centering
    \resizebox{0.7\linewidth}{!}{
    \begin{tabular}{ cccccccc } 
        \toprule
        $N_{d}$ & $N_{o}$ & Inter & Dist F1 & Occl F1 & Dist p / r & Occl p / r\\ 
        \midrule
        1 & 1 & F & 36.3  & 38.3 & 36.6 / 36.0 & 38.6 / 38.0\\
        1 & 1 & \textbf{T} & \textbf{36.5} & \textbf{39.0} & \textbf{36.9} / \textbf{36.1} & \textbf{39.4} / \textbf{38.6} \\\midrule
        3 & 3 & F & 38.1 & 40.5 & 38.5 / 37.7 & 41.0 / 40.1\\
        3 & 3 & \textbf{T} & \textbf{38.6} & \textbf{41.2} & \textbf{38.9} / \textbf{38.2} & \textbf{41.6} / \textbf{40.9}\\\midrule
        6 & 6 & F & 38.0 & 40.3 & \textbf{38.5} / 37.6 & 40.8 / 39.9\\ 
        6 & 6 & \textbf{T} & \textbf{38.1} & \textbf{40.8} & 38.4 / \textbf{37.8} & \textbf{41.2} / \textbf{40.5}\\\bottomrule
    \end{tabular}
    }
    \label{effects_of_intersection}
\end{table}

\subsubsection{Predict the Generalized Intersection Box when No Intersection Exists}\label{sec:experiment_predicte_intersection_only_when_intersection_exists}
\textcolor[RGB]{0,0,0}{We investigate the function of predicting GIT when no intersection exists for a pair of objects (\cref{git_when_no_intersection}). For models trained with GIT, we use the best setting found in previous experiments: 6, 3, and 3 decoder layers in object pair decoder, distances decoder, and occlusion decoder, respectively. Additionally, since generalized intersection boxes are not always predicted, the matcher does not consider the costs of generalized intersection boxes. }

\textcolor[RGB]{0,0,0}{When trained to predict the intersection region if and only if an intersection exists, we observe a 2\% and a 2\% relative F1-score drop on distance and occlusion, respectively. Additionally, we observe no performance boost compared to the model trained without GIT if the model does not predict the generalized intersection boxes when no intersection exists.}

Additionally, for the ``no occlusion" class, we observe a 3\% relative F1-score drop if the model does not predict the intersection box when no intersection exists.

Predicting the generalized intersection box even when no intersection exits is critical for our model's learning of relative occlusion relationship detection. It guides our model to identify the region where intersections may or may not happen. Specifically, when no intersection exists, it still guides our model to identify a region, which can be used to conclude no occlusion exists. In other words, to conclude no intersection exists, our model still needs to know where to look at. Not knowing where to look would degrade the performance when detecting ``no occlusion". Our proposed GIT consistently helps the model to identify the region that is most critical for the conclusion of the relative occlusion relationships no matter intersection exists or not.

\begin{table}[t]
    \caption{\textcolor[RGB]{0,0,0}{Effects of predicting the generalized intersection box when no intersection exists. (PINI: predict the generalized intersection box when no intersection exists.  Occl: no occl F1: the F1 score for the class ``no occlusion")}}
    \centering
    \resizebox{0.85\linewidth}{!}{
    \begin{tabular}{ cclllll } 
        \toprule
        Inter & PINI & Dist F1 & Occl F1 & Occl: no occl F1 & Dist p / r & Occl p / r\\ 
        \midrule
        F & - & 38.1   &  40.5   & 41.6  & 38.5 / 37.7 & 41.0 / 40.1 \\
        T & F & 37.9 \textcolor[RGB]{0,200,0}{(-2\%)} &  40.5  \textcolor[RGB]{0,200,0}{(-2\%)} & 41.3 \textcolor[RGB]{0,200,0}{(-3\%)} & 38.3 / 37.5 & 41.0 / 40.0 \\
        T & \textbf{T} & \textbf{38.6}  &  \textbf{41.2} & \textbf{42.4} &  \textbf{38.9} / \textbf{38.2} & \textbf{41.6} / \textbf{40.9}\\
        \bottomrule
    \end{tabular}
    }
    \label{git_when_no_intersection}
\end{table}

\begin{table}[t]
    \caption{\textcolor[RGB]{0,0,0}{Effects of not using a dedicated decoder for object pair detection. (DOPD: dedicated object pair decoder. In other words, using one decoder for all three sub-tasks.)}}
    \centering
    \resizebox{0.7\linewidth}{!}{
    \begin{tabular}{ cllll } 
        \toprule
        DOPD & Dist F1 & Occl F1 & Dist p / r & Occl p / r\\ 
        \midrule
        F & 37.1  \textcolor[RGB]{0,200,0}{(-4\%)} &  40.0 \textcolor[RGB]{0,200,0}{(-3\%)} & 37.4 / 36.8 & 40.3 / 39.6 \\
        \textbf{T} & \textbf{38.6}  &  \textbf{41.2} & \textbf{38.9} / \textbf{38.2} & \textbf{41.6} / \textbf{40.9}\\
        \bottomrule
    \end{tabular}
    }
    \label{single_decoder}
\end{table}

\subsubsection{Separate Decoders}

\textcolor[RGB]{0,0,0}{By performing experiments without a dedicated object pair decoder, we study how using separate decoders benefits the learning of relationships. Specifically, we use only one decoder with six layers to detect object pairs, relative distances, and relative occlusions. We observe a 4\% and a 3\% relative performance drop in relative distance and relative occlusion detection, respectively (\cref{single_decoder}). }

\textcolor[RGB]{0,0,0}{The drops in performance indicate that using separate decoders is conducive to the learning of relative distance and relative occlusion relationships. Specifically, using a dedicated object pair decoder frees the two relationship decoders from detecting objects, allowing them to focus on task-specific regions. At the same time, using two relationship decoders for two different types of relationships helps each decoder to focus on its own task-specific without distracting each other. }

\textcolor[RGB]{0,0,0}{We provide attention weight visualizations and analysis to demonstrate the benefits of using separate decoders in \cref{sec:visualizations_cascade_vs_single} and \cref{sec:analyze_effects_of_using_separate_decoders}, respectively.}

\subsection{Qualitative Results}
\begin{figure}[t]
  \centering
  \includegraphics[width=0.8\linewidth]{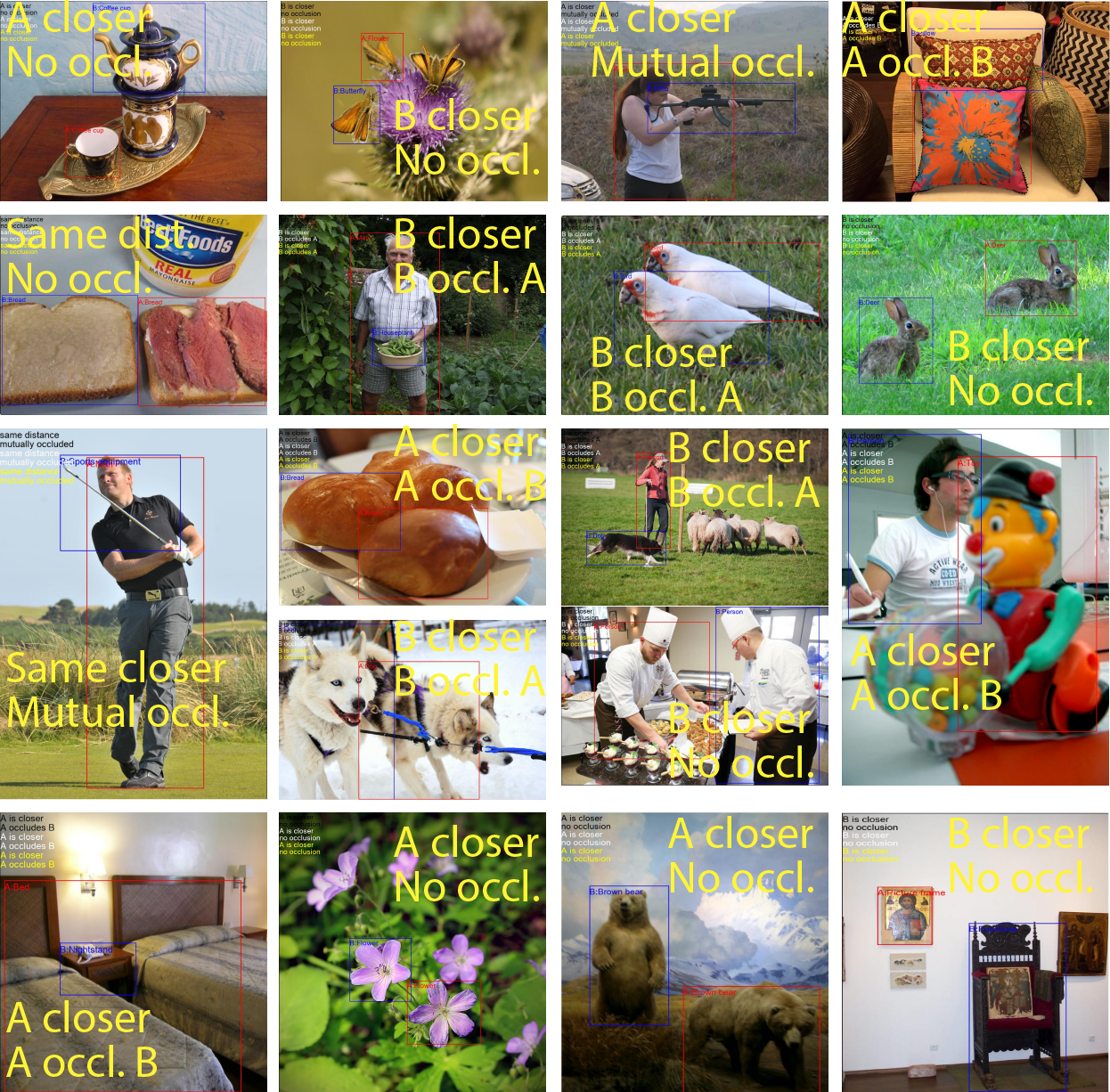}
  \caption{Qualitative examples: correct predictions.}
  \label{qualitative_correct} 
\end{figure}

\begin{figure}[t]
  \centering
  \includegraphics[width=0.8\linewidth]{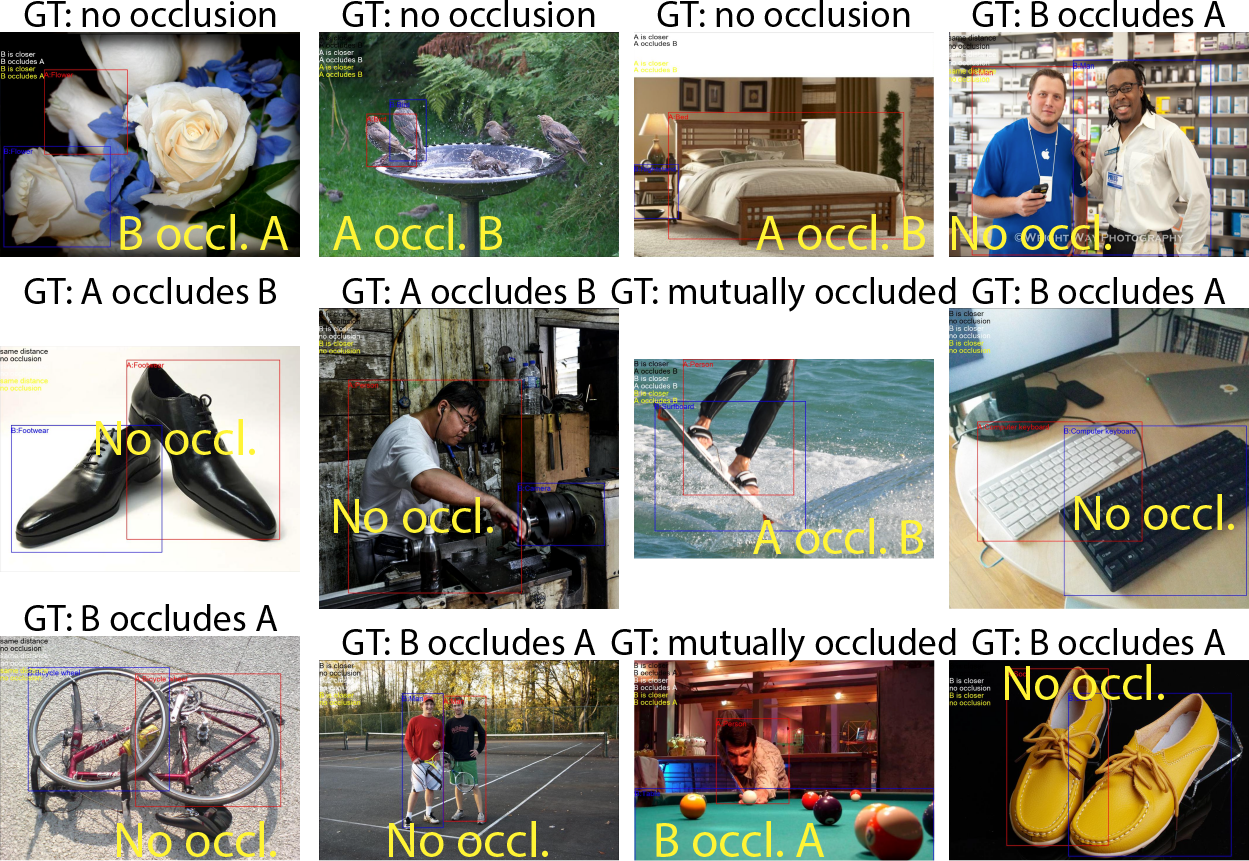}
  \caption{Qualitative examples: incorrect relative occlusion predictions.}
  \label{qualitative_incorrect_occlusion} 
\end{figure}

\begin{figure}[t]
  \centering
  \includegraphics[width=0.8\linewidth]{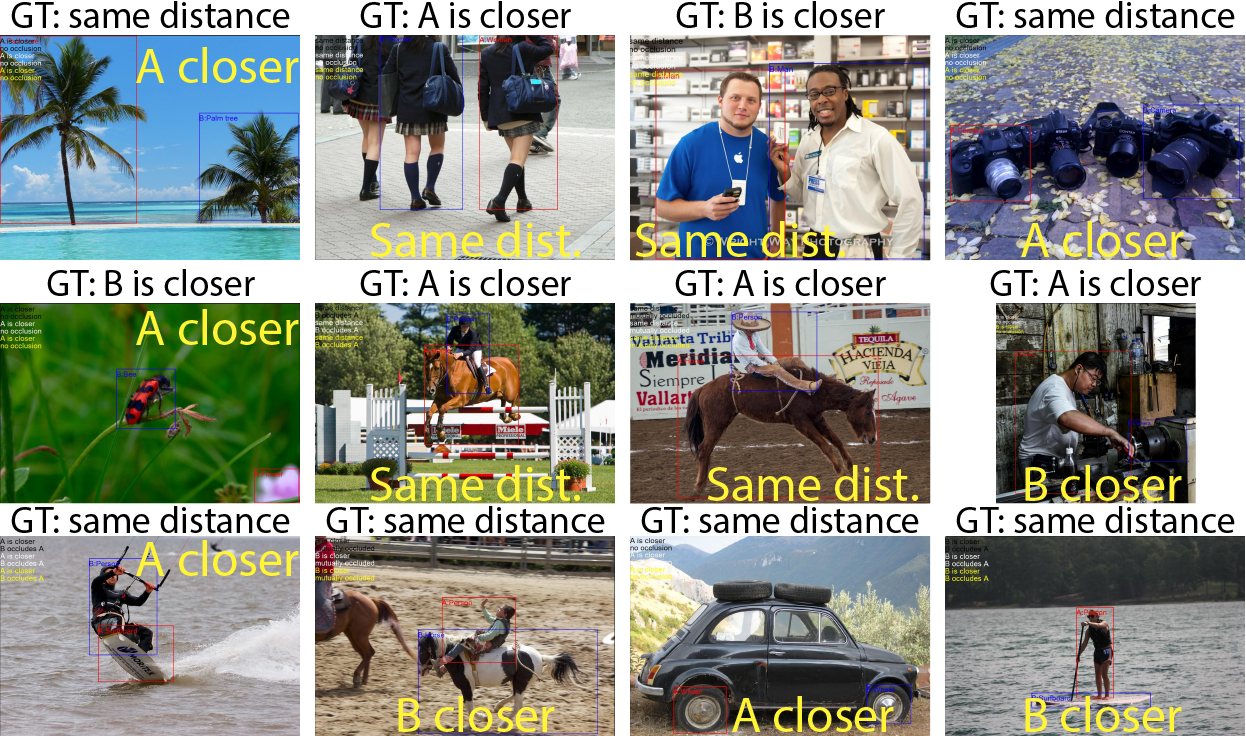}
  \caption{Qualitative examples: incorrect relative distance predictions.}
  \label{qualitative_incorrect_distance} 
\end{figure}

\begin{figure}[t]
  \centering
  \includegraphics[width=0.8\linewidth]{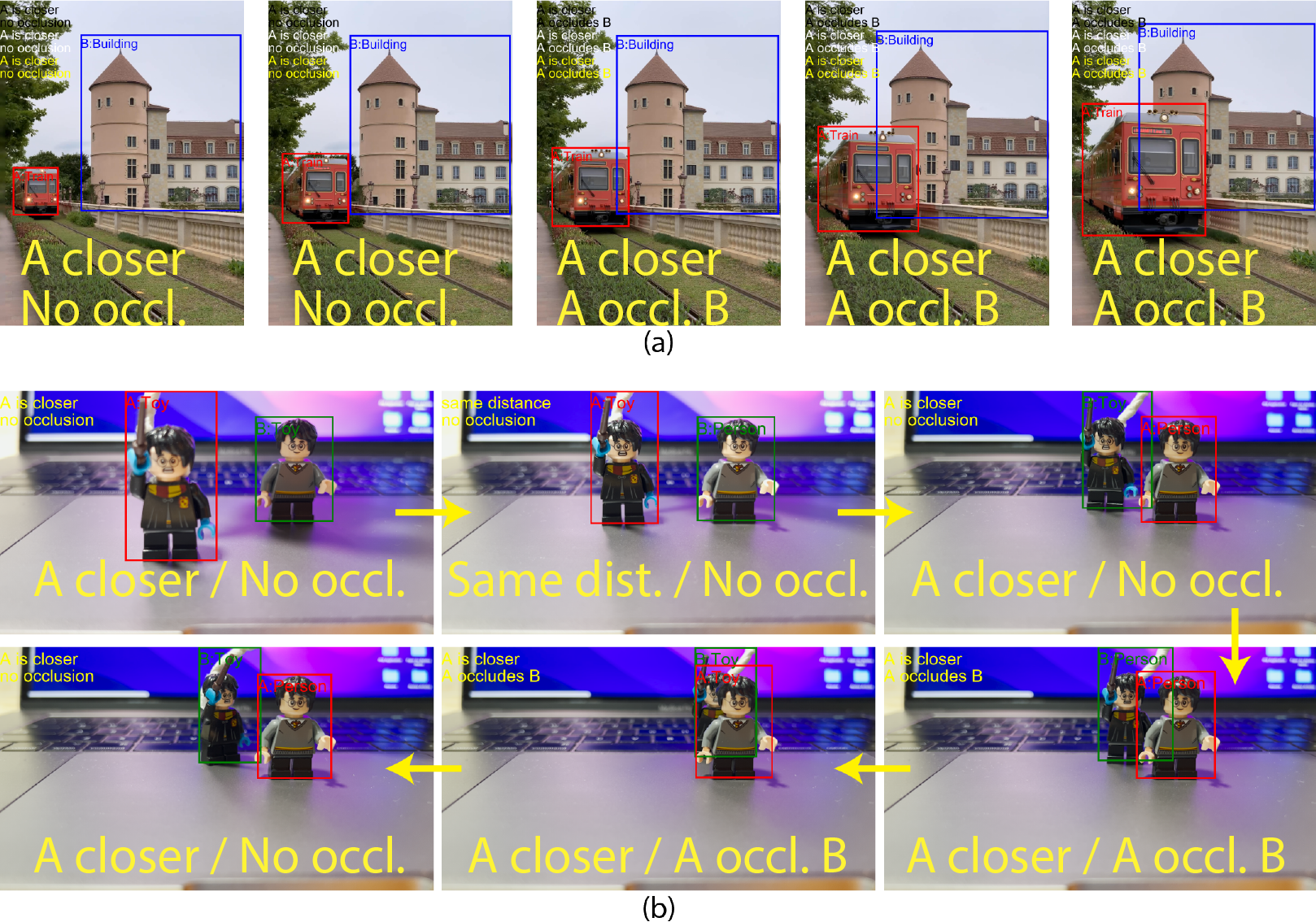}
  \caption{Model performance on videos. (a) Change in relative occlusion relationship: a moving train gradually occludes the building. (b) Change in both relative distance and relative occlusion relationships: a toy (on the left) moves further away from the viewpoint, moves right until being occluded by the other toy, and moves left and becomes not occluded. Links to videos are in the footnote on the first page.}
  \label{demo_video_frames} 
\end{figure}

On the 2.5VRD dataset, qualitative results of our model are given in \cref{qualitative_correct} (correct predictions), \cref{qualitative_incorrect_occlusion} (incorrect occlusion), and \cref{qualitative_incorrect_distance} (incorrect distance). Qualitative results on these images demonstrate that our model correctly predicts relative distance and occlusion relationships in diverse scenarios but fails under some situations. We hope these failure cases can inspire readers to further improve model performance on this fundamental task.

We also run our model \textcolor[RGB]{0,0,0}{(trained on the 2.5VRD dataset without any extra data)} on videos that contain moving objects with changing relationships. These videos are provided by authors and are not from the 2.5VRD dataset. Links to the videos are given in the footnotes on the first page. 

\textcolor[RGB]{0,0,0}{At the upper left corner of each figure, texts are printed in three different colors (black, white, and yellow). Except for the color, they are the same. }

\subsection{Attention Weight Visualizations}
\begin{figure*}[t]
  \centering
  \includegraphics[width=1\linewidth]{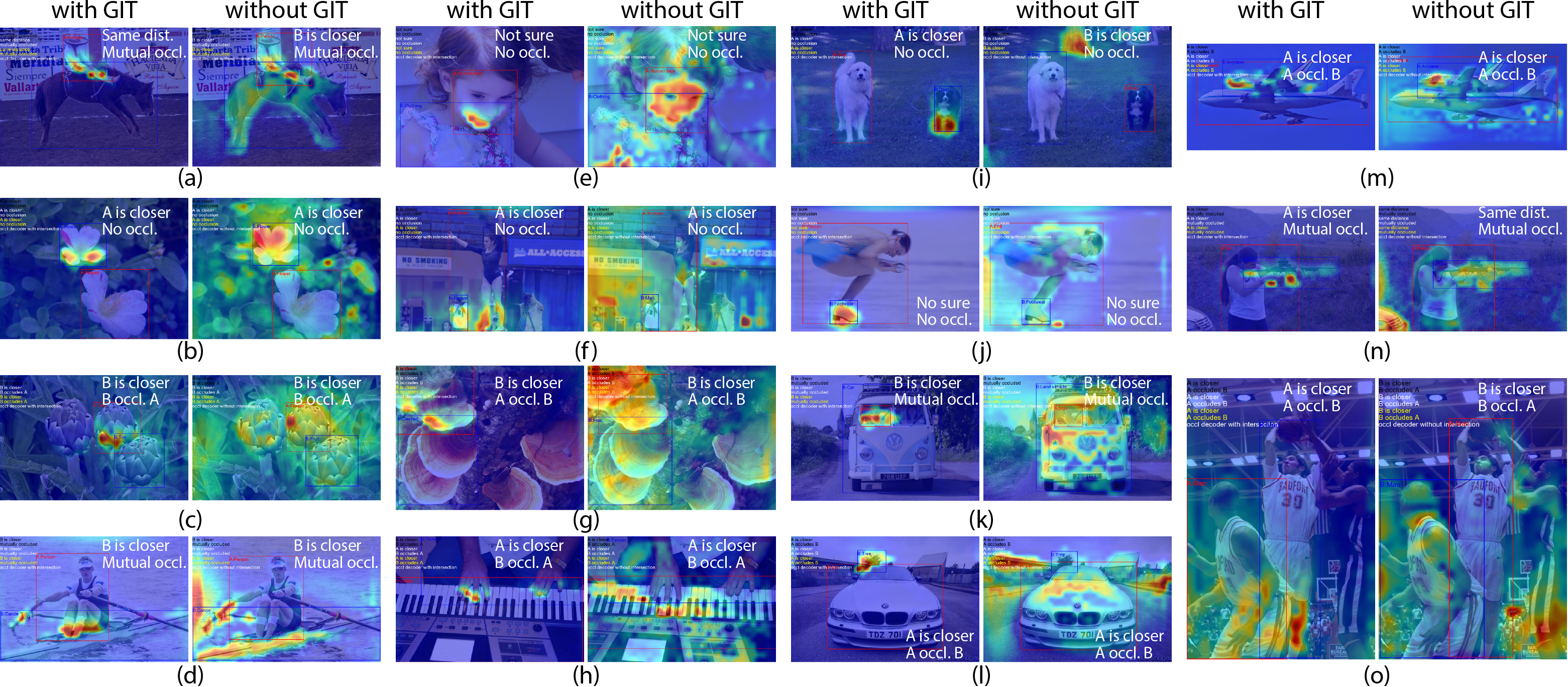}
  \caption{Visualizations of attention weights in our occlusion decoders. The generalized intersection box prediction task (GIT) effectively guides our model to focus on occlusion-specific regions when detecting relative occlusion relationships. (Red indicates heavy attention. Blue indicates nearly no attention.)}
  \label{attention_easy} 
\end{figure*}

\begin{figure}[t]
  \centering
  \includegraphics[width=1\linewidth]{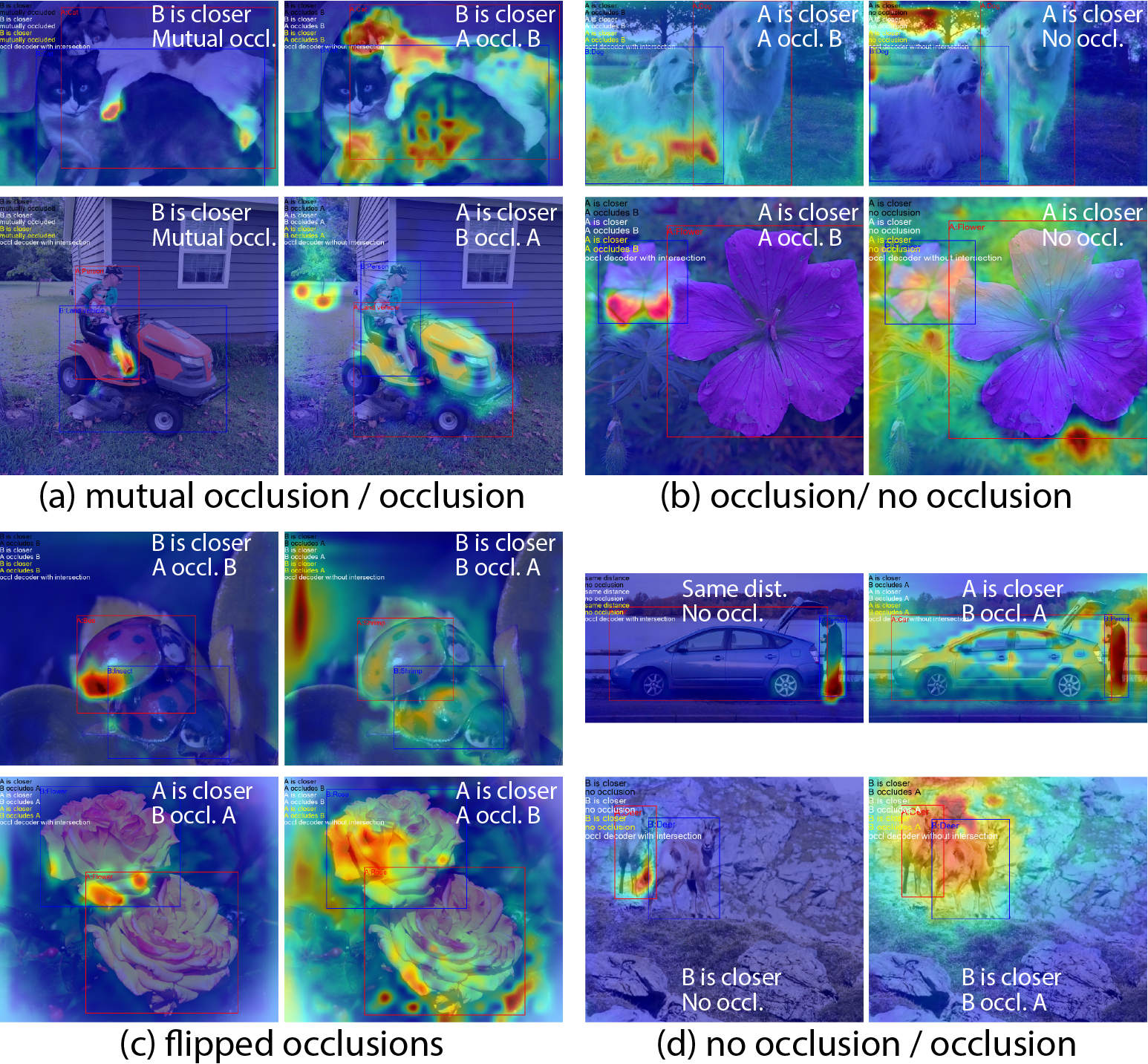}
  \caption{Improvements in relative occlusion detection (from incorrect to correct) with the generalized intersection box prediction task (GIT). (Left: with intersection box prediction task. Right: without intersection box prediction task)}
  \label{improvements_with_intersection}
\end{figure}

We provide visualizations of attention weights of our decoders. We compare and contrast the attention weights of occlusion decoders trained with and without GIT. The occlusion decoder trained with GIT demonstrates much more focused attention (\cref{attention_easy}). In some images, the occlusion decoder trained with GIT exhibits improved prediction correctness (\cref{improvements_with_intersection}). Additionally, we compare visualizations of all three decoders in multi-decoder architecture with those of the decoder in the single-decoder architecture. We observe that using separate decoders for different sub-tasks allows each decoder to focus on the most relevant information.

\begin{figure}[t]
  \centering
  \includegraphics[width=0.9\linewidth]{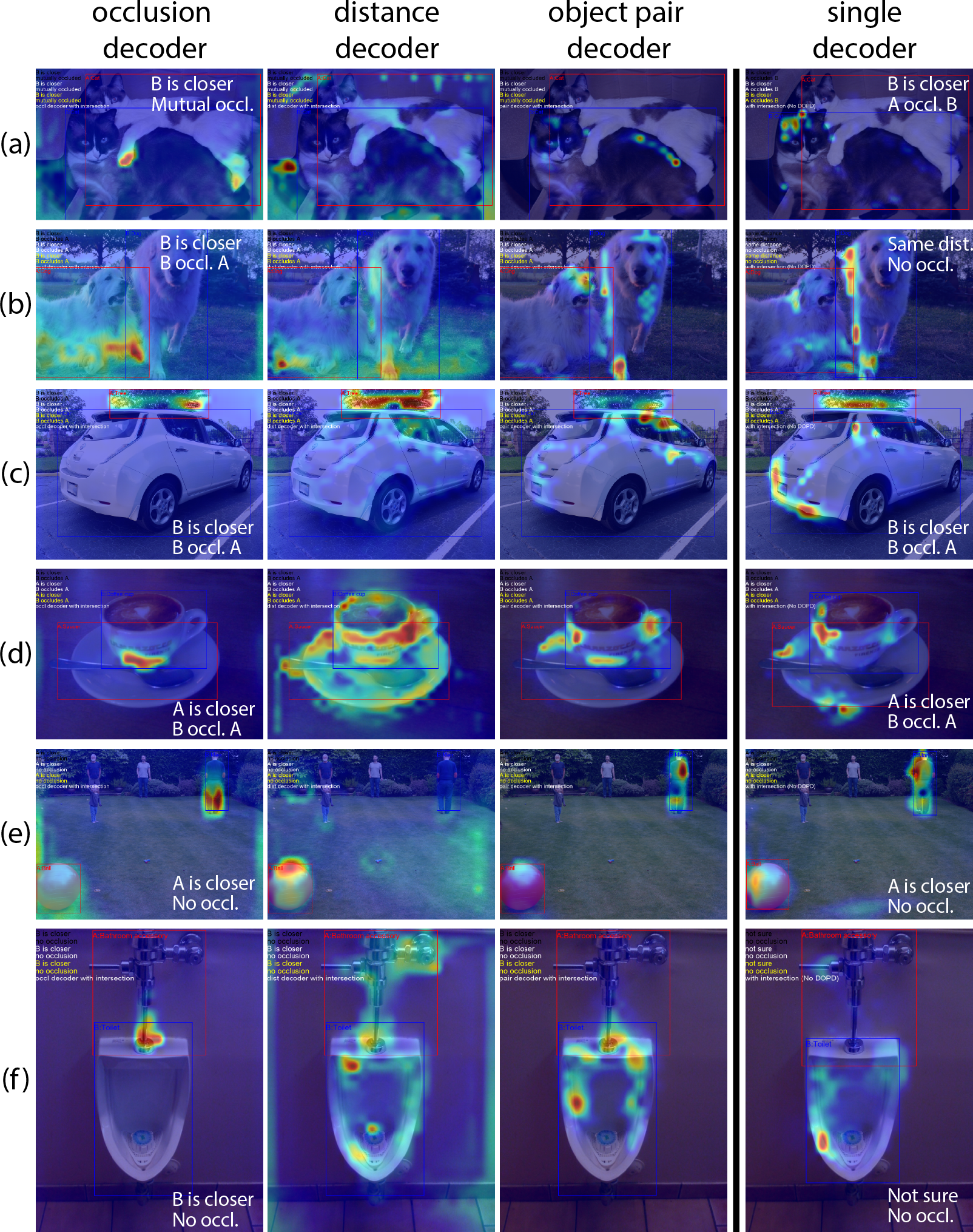}
  \caption{Attention weights in each decoder when using separate decoders (left) or in the only decoder when not using separate decoders (right). Using separate decoders allows the model to focus on most-relevant information when addressing each sub-task.}
  \label{cascade_vs_single}
\end{figure}

\subsubsection{GIT: More focused attention}
With GIT, our occlusion decoder shows more focused attention to areas around occlusion sites (mostly when occlusions exist) or object extreme points close to another object (mostly when no occlusion exists) and focus much less on irrelevant areas such as regions outside the union of the object pair (\cref{attention_easy}). 

For example, in \cref{attention_easy} (b) left where no occlusion exists, our occlusion decoder trained with GIT pays heavy and focused attention to the bottom of the upper flower, which is the extreme point of the upper flower that is the closest to the lower flower. In contrast, in \cref{attention_easy} (b) right, the one trained without GIT pays heavy attention to the upper left portion of the upper flower, which is very far away from the lower flower and is very unlikely to be a site of occlusion. 

When occlusion does exist (\cref{attention_easy} (c)), the occlusion decoder trained with GIT attends heavily only to regions between the pair of flowers but the occlusion decoder trained without GIT attends heavily to multiple locations outside the pair of flowers. Similarly, in \cref{attention_easy} (l), the occlusion decoder trained with GIT pays focused attention to the region between the car and the tree. In contrast, the occlusion decoder trained without GIT attends heavily to the rooftop and fence on both sides of the photo (\cref{attention_easy} (l)). In the rest of the figures in \cref{attention_easy}, similar patterns can also be observed.

\subsubsection{GIT: Improvements in correctness}
In \cref{improvements_with_intersection} (a), the occlusion decoder trained with GIT focuses much more on the areas that are mutually occluded and attends much less to areas that are outside the union of the object pair. In specific, in the upper two images of \cref{improvements_with_intersection} (a), the occlusion decoder trained with GIT focuses much more on the white cat's front paw (which occludes the black cat) and hind paw (which is occluded by the black cat). In the lower two images of \cref{improvements_with_intersection} (a), the occlusion decoder trained with GIT focuses much more on areas near the man's leg (which occludes the vehicle) and feet (which is partially occluded by the vehicle).

In addition to having more focused attention, model trained with GIT exhibits improved prediction correctness. Specifically, in \cref{improvements_with_intersection} (a), the model trained with GIT correctly predicts that mutual occlusions exist between each of the two predicted pairs of objects. In contrast, the model trained without GIT only predicts that one object occludes the other object. 

Similarly, in \cref{improvements_with_intersection} (b), the occlusion decoder trained with GIT pays more focused attention to the front left leg of the dog on the left (in the upper image) and the bottom portion of the upper flower (in the lower image) and effectively suppresses attention to the tree (in the upper image) and the surrounding leaves (in the lower image). In these two pairs of images, the model trained with GIT successfully detects occlusion while the model trained without GIT incorrectly predicts that no occlusion exists. 

Furthermore, in \cref{improvements_with_intersection} (d), the occlusion decoder trained with GIT attends heavily to the areas between the objects which are very close to each other and successfully predicts that no occlusion exists. The occlusion decoder trained without GIT, in contrast, fails to conclude no occlusion exists between the two pairs of objects in \cref{improvements_with_intersection} (d).

\subsubsection{Separate Decoders}\label{sec:visualizations_cascade_vs_single}

We provide the attention weight visualizations of occlusion decoder, distance decoder, and object pair decoder for our proposed multi-decoder model (left 3 columns of \cref{cascade_vs_single}). We also provide the attention weights for the single-decoder model (the rightmost column of \cref{cascade_vs_single}). Both model are trained with the GIT.

In the following section, We will provide analyses of attention weight visualizations.

\section{Analyses}
In this section, we provide analyses for the observed more focused attention and improvements in prediction correctness in the model trained with the generalized intersection box prediction task (GIT). We also analyze the differences between decoder attention visualizations for the multi-decoder model and the single-decoder model. Additionally, we analyze the effects of object size and location. We also study the source of errors. Finally, we compare state-of-the-art VRD methods with our proposed method.

\subsection{Attention Weight Analysis}
\subsubsection{GIT}
Visualizations of attention weights in the experiment section demonstrate that GIT can lead to more focused attention and improvements in prediction correctness. We provide analyses for these observed results in this subsection. 

GIT effectively guides attentions to locations near occlusions because the generalized intersection box proposed in \cref{intersection} usually encloses the actual occlusion sites when there are occlusions. For example, in \cref{demo_video_frames} (a), the train gradually moves to the right and occludes the building. Starting from the third frame in \cref{demo_video_frames} (a), the train occludes the building, and the intersection happens inside the intersection of the red and blue bounding boxes. Since occlusions usually happen inside the intersection box, asking the occlusion decoder to predict intersection boxes effectively guides the occlusion decoder to attend to areas near occlusions.

Guiding the occlusion decoder's attention to the intersection box helps the decoder to focus on actual occlusions and protects the occlusion decoder from distractions. In \cref{improvements_with_intersection} (a), the occlusion decoder trained with GIT demonstrated much more focused attention to the paws of the white cat (in the upper image) and the areas between the man's leg and the vehicle (in the lower image). At the same time, occlusion's attention to irrelevant regions was greatly reduced. In specific, in the upper two images of \cref{improvements_with_intersection} (a), the occlusion decoder trained with GIT effectively suppressed its attention to the legs and belly of the black cat and the back of the white cat; in the lower two images of \cref{improvements_with_intersection} (a), the occlusion decoder trained with GIT effectively ignored the tree far away from the object pair and focused on areas near the man's feet instead of the anterior and lateral parts of the vehicle. 

Paying increased attention to regions near actual occlusion sites helped the decode to detect more occlusion sites inside the intersection box. At the same time, ignoring distractions from irrelevant regions helped the occlusion decoder to make decisions using information relevant to occlusion relationships. Much more focused attention to relevant regions and significantly decreased attention to distracting regions might contribute to the improvements in relative occlusion relationship predictions in these images.

When no occlusion exits, predicting the generalized intersection box is still very beneficial because attending to object parts near the intersection box is important for concluding that no occlusion exists, especially when two objects are very close to each other. For example, in \cref{improvements_with_intersection} (d), the car is very close to the human and the two dears are very close to each other. To determine whether relative occlusions exist or not, paying close attention to regions near the tiny space between the car and the human and between two dears is required. Specifically, at first glance, the only potential occlusion site of the two dears in \cref{improvements_with_intersection} (d) is the region between the front left leg of the left dear and the hind right leg of the right dear. Paying closer attention to this potential occlusion site would lead to the conclusion that two dears do not occlude each other. 

Paying focused attention to potential occlusion sites was exactly what the occlusion decoder did when determining the relative occlusion relationships between the two dears. Specifically, in \cref{improvements_with_intersection} (d), the occlusion decoder trained with GIT allocated heavy and focused attention to the regions near the front left leg of the left dear and correctly concluded no occlusion exists between two dears. The occlusion decoder trained without GIT, in contrast, paid much less attention to this potential occlusion site. Instead, it allocated heavy attention to the head of the left dear and the rocks above dears. Failing to pay focused attention to the potential occlusion site and being distracted by irrelevant information help to explain why the occlusion decoder trained without GIT failed to conclude that no occlusion exists between the two dears.

\subsubsection{Separate Decoders}\label{sec:analyze_effects_of_using_separate_decoders}

\paragraph{The most relevant information}
\textcolor[RGB]{0,0,0}{Using separate decoders for occlusion, distance, and the object pair allows each decoder to focus on the most relevant information. }

\textcolor[RGB]{0,0,0}{For example, in \cref{cascade_vs_single} (a), the occlusion decoder attends heavily to the white cat's front paw (which occludes the black cat) and hind paw (which is occluded by the black cat). Attending to these paws is very helpful for concluding mutual relationships. The distance decoder mainly attends to the ground, wall, and the cats' contours. Attending to these regions helps the model to understand the spatial configurations. The object pair decoder attends heavily to the black cat's head, chest,  back, and tail. Identifying these representative parts helps the object pair decoder recognize the cat. }

\paragraph{Omission of important details}
\textcolor[RGB]{0,0,0}{In contrast, when using a single decoder for all three sub-tasks, the decoder has to attend to much more information. The requirement to attend to too much information makes focusing on the most-relevant part more challenging. This increased difficulty can lead to the omission of some important details, making it more difficult to correctly detect occlusion relationships. }

\textcolor[RGB]{0,0,0}{For instance, in the rightmost column of \cref{cascade_vs_single} (a), the single decoder attends extensively to the cats' bodies, with a greater emphasis near the black cat's ear, white cat's wrist, and white cat's chest. The decoder's attention to the white cat's chest and wrist could only indicate that the white cat is occluding the black cat. This single decoder, unfortunately, paid only a small amount of attention to the white cat's front and hind paws. Failing to pay enough attention to the white cat's hind paw makes it much harder to identify that the white cat is occluded by the black cat. This omission of this critical detail partly explains why the single decoder fails to detect mutual occlusion. Similarly, in \cref{cascade_vs_single} (b) the single decoder fails to pay sufficient attention to the left front paw of the dog on the left side (in the red box). This paw is the only occlusion site between these two dogs. Failing to attend to this paw makes it unlikely to detect the occlusion relationship between these two dogs. The omission of the occlusion of the paw helps to explain why the single decoder fails to detect occlusion between the two dogs. }

\paragraph{Distracting information}
\textcolor[RGB]{0,0,0}{At the same time, using a single decoder could introduce distracting information for relationship detection. }

\textcolor[RGB]{0,0,0}{For instance, in \cref{cascade_vs_single} (a), the single decoder's intense attention on the black cat's ear introduce information that is irrelevant to the detection of mutual occlusion. In specific, the black cat's ear (upper left corner of the black cat's head) does not occlude the white cat. This information could only help the detection of ``no occlusion." Detecting that the black cat's ear does not occlude the white cat obviously does not contribute to the detection of mutual occlusion relationship. Therefore, this information (the black cat's ear) acts as distractions for relative occlusion relationship detection. Similarly, in \cref{cascade_vs_single} (c), the only occlusion is located at the intersection of the car and the tree. Information fron the back of the car is distracting and does not help the occlusion relationship detection. }

\paragraph{Reduce distractions}
\textcolor[RGB]{0,0,0}{Using a separate object pair decoder helps to partially eliminate these distracting information when detecting relative occlusion relationships. }

\textcolor[RGB]{0,0,0}{For instance, the specialized occlusion decoder does not attend to the black cat's ear in \cref{cascade_vs_single} (a), the back of the car in \cref{cascade_vs_single} (c), the edge of the saucer in \cref{cascade_vs_single} (d), and the lower half of the toilet in \cref{cascade_vs_single} (f). Instead, the specialized occlusion decoder achieves highly focused attention to the regions where occlusions take place in these images. }

\subsection{Object Size and Location Analysis}
\begin{figure}[t]
  \centering
  \includegraphics[width=1.00\linewidth]{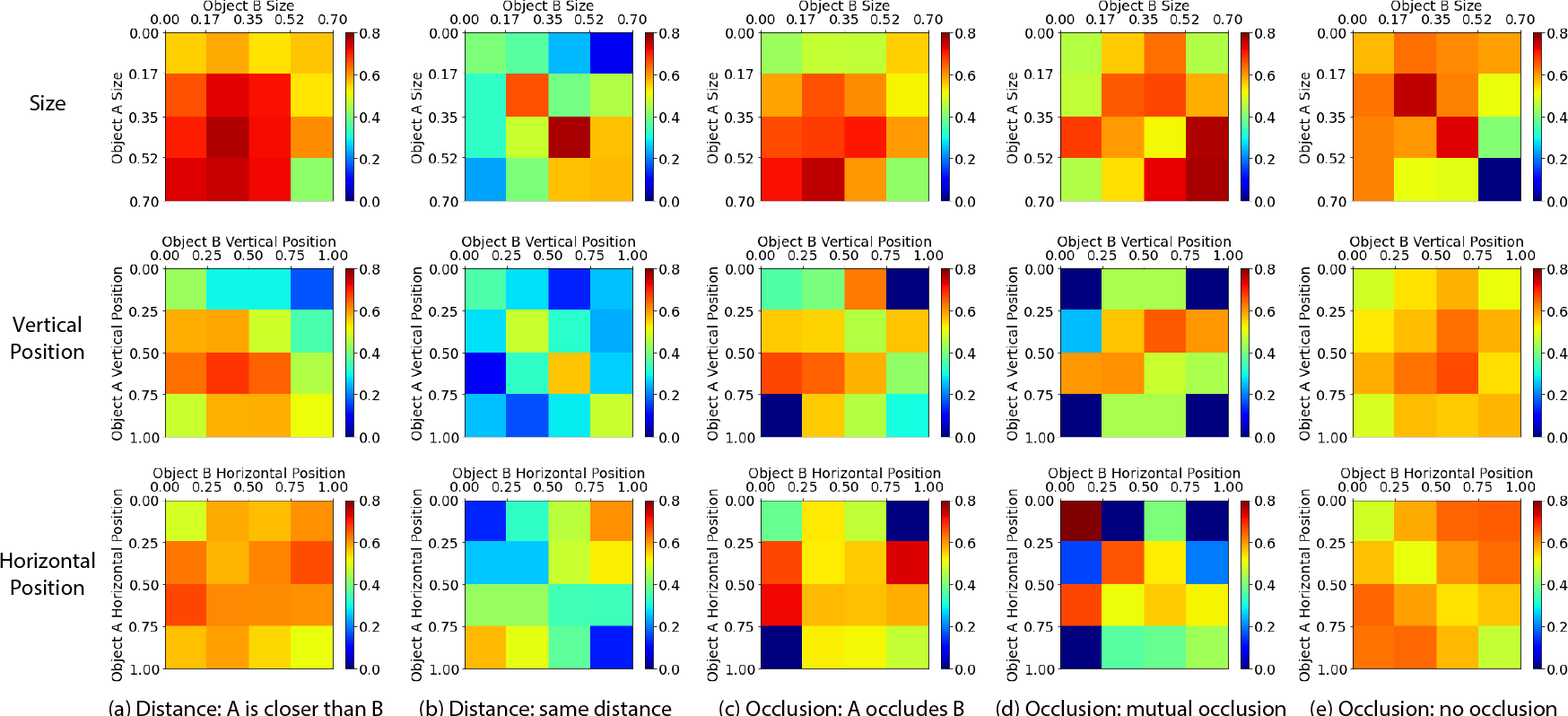}
  \caption{Model performance (F1-scores) w.r.t object size, vertical position, and horizontal position. }
  \label{size_and_position} 
\end{figure}
We study the effects of object sizes, vertical positions, and horizontal positions to explore how our model predicts the relative relationships leveraging these priors. F1-score distribution maps are shown in \cref{size_and_position}.

An object looks larger in the photo as it gets closer to the viewpoint. Leveraging this prior distribution in natural images, the model can learn larger objects are more likely to be closer to the viewpoint while smaller objects are more likely to be further. As a result, correctly concluding that object A is closer to the viewpoint when object A is very small or object B is very large can be more challenging for the model. For the relationship ``A is closer than B", F1-scores are low when object A size is small and when object B size is large (\cref{size_and_position} (a) upper). For the same reason, F1-scores are relatively high when two objects are of similar sizes when determining ``same distance" (\cref{size_and_position} (b) upper figure). 

Another prior in natural images is that, for objects on the ground, those whose base is further to the horizon are closer than those whose base is closer to the horizon. In other words, objects on the ground are more likely to be further away if their vertical location is high. Consequently, it is rare for object A to be closer than object B when the vertical position of object A in the image is very high while that of object B is very low. In \cref{size_and_position} (a) second row, the lower F1 score part shows that it is more difficult for our model to predict the ``A is closer than B" results correctly when A is very high or B is very low.  And F1-score in the first column gets lower from left to right as the location of object B gets lower. This indicates that our model does consider the vertical location of objects when deciding the relative distance relationships. 



For relationships in which A and B are interchangeable (e.g., `A and B have same distance' is equal to `B and A have same distance'), there is no particular preference for either the size or position of A or B. Therefore, the size and position distributions of A and B are similar, which is reflected in \cref{size_and_position} (b) (d) (e), that the F1-score distribution maps are relatively symmetric about their principal diagonals.

\subsection{Error Source Analysis}
As an end-to-end model of relationship detection, the overall performance of our model is determined by two major factors: object pair proposal and relationship classification. To study the performance of our relationship classifier, we attempt to bypass the effects of detection error. Since it is virtually impossible to obtain a perfect object detection (predicted bounding boxes are identical to ground truth), we need to use the notion of ``good detection" to filter our predictions. An IoU threshold of $0.6$ is used to select predictions and we computed the precision of occlusion/distance classification within this group, see \cref{tab:Almost_Perfect_Precision}. From the table, we can see that the model with GIT outperforms the model without GIT when an occlusion exists (``A occludes B", ``B occludes A", or ``mutual occlusion"). In general, for both models, the relationship detection precision for ``good predictions" is much higher than their overall F1-scores. This result shows that the main bottleneck is to devise a better object detection model so that most of the object pairs can be proposed.

\begin{table}[t]
\centering
\caption{Prediction Precision in good detections}
\resizebox{0.6\linewidth}{!}{
\begin{tabular}{lll}
\toprule
Relationship Class & Without GIT & With GIT \\
\midrule
No occlusion                & 0.96                 & 0.95              \\
A occludes B                & 0.74                 & 0.78  \textcolor[RGB]{205,0,0}{(+5\%)}             \\
B occludes A                & 0.73                 & 0.76  \textcolor[RGB]{205,0,0}{(+4\%)}           \\
Mutual occlusion            & 0.68                 & 0.69  \textcolor[RGB]{205,0,0}{(+1\%)}           \\
Occlusion total             & 0.90                 & 0.91              \\
\midrule
Distance not sure           & 0.90                 & 0.92              \\
A is closer                 & 0.87                 & 0.86              \\
B is closer                 & 0.87                 & 0.86              \\
Same distance               & 0.62                 & 0.64              \\
Distance total              & 0.85                 & 0.85 \\  
\bottomrule
\label{tab:Almost_Perfect_Precision}
\end{tabular}
}
\end{table}

\subsection{Compare with VRD methods}
\begin{figure}[t]
  \centering
  \includegraphics[width=1\linewidth]{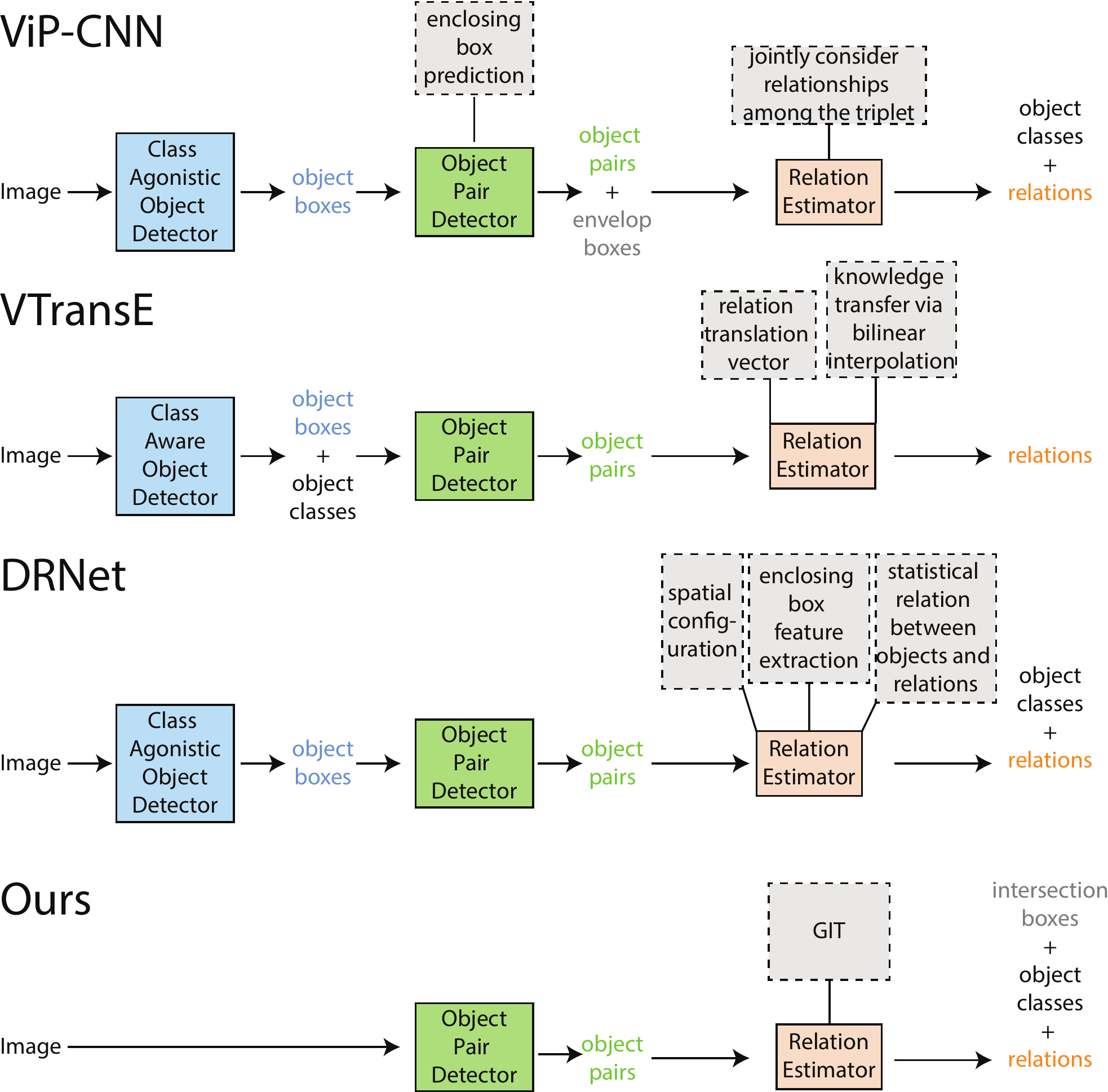}
  \caption{\textcolor[RGB]{0,0,0}{Compare ViP-CNN, VTransE, and DRNet with our method.}}
  \label{fig:compare_with_vrd} 
\end{figure}

\textcolor[RGB]{0,0,0}{We compare VRD methods, DRNet~\citep{DRNet}, ViP-CNN~\citep{ViP-CNN}, VTransE~\citep{VTransE}, and PPR-FCN~\citep{PPR-FCN}, with ours (\cref{fig:compare_with_vrd}}). 

\paragraph{DRNet}
\textcolor[RGB]{0,0,0}{DRNet \cite{DRNet}, as shown in the third row of \cref{fig:compare_with_vrd}), exploits statistical relation between objects and relations and enclosing boxes (a box that contains both object A and object B ``with a small margin"). Specifically, DRNet models and refines the estimated probability between objects and relationships. For example, it aims to assign a higher probability to reasonable ``(cat, eat, ﬁsh)" than impossible ``(ﬁsh, eat, cat)" \cite{DRNet}. Additionally, it extracts features for the enclosing boxes to obtain contextual information that helps decide the relationships between a pair of objects \cite{DRNet}. }

\textcolor[RGB]{0,0,0}{While the above two types of information are useful for more general and high-level relationship detection tasks such as VRD and HOI detection, they are much less helpful when detecting relative occlusion and relative distance relationships. Take the cat-eat-fish example given by \cite{DRNet} as an example, cat is closer than fish and fish is closer than cat are both reasonable predictions. Similarly, a cat can occlude fish and a fish can occlude fish. In other words, when detecting relative distance and relative occlusion relationships, the statistical relationships between objects and relationships are less strong and less useful for relative distance and occlusion relationship detection.}

\textcolor[RGB]{0,0,0}{Additionally, to detect the relative occlusion relationships, the model only needs to attend to the generalized intersection box instead of the larger enclosing box. Compared to our proposed generalized intersection box, the larger enclosing box used by DRNet contains too much information that is not germane to relative occlusion relationship detection. The information that is most relevant to relative occlusion relationship detection is located in the region where two bounding boxes intersect because occlusion cannot happen outside of this region. Thus, features extracted by DRNet for the enclosing box are too redundant. In contrast, our proposed GIT helps our model learn to focus on the potential occlusion sites (regions where two bounding boxes intersect) when detecting relative occlusion relationships. In contrast, our proposed generalized intersection box provides more relevant information for relative occlusion relationship detection.}

\paragraph{ViP-CNN}
\textcolor[RGB]{0,0,0}{ViP-CNN \cite{ViP-CNN}, which predicts an enclosing box that ``tightly covers both" objects similar to the one in DRNet (the first row of \cref{fig:compare_with_vrd}), similarly suffers from the redundancy problem as DRNet.}

\paragraph{VTransE}
\textcolor[RGB]{0,0,0}{VTransE considers relationship detection ``as a vector translation" such that, for the features in a low-dimensional space, object A + relation $\approx$ object B~\citep{VTransE}. By learning the vector translation (the second row of \cref{fig:compare_with_vrd}), VTransE aims to make the learning of the triplet (object A, relation, B) less challenging ``by avoiding learning the diverse appearances" of the triplet ``with large variance"~\citep{VTransE}. While its strategy makes easier the learning of more general relationships in which the detailed appearances of objects contribute less to the relationship, this strategy makes the learning of relative occlusion relationships very challenging because learning the relative occlusion requires learning the subtle details of object appearances, such as which parts are occluded. Therefore, VTransE's strategy to avoid learning diverse appearances partly contributes to its sub-optimal performance when detecting relative occlusion relationships.}

\paragraph{PPR-FCN}
\textcolor[RGB]{0,0,0}{PPR-FCN~\citep{PPR-FCN} is designed for weakly supervised VRD (thus not inlucded in the comparisons of \cref{fig:compare_with_vrd}), where the ground truth is image level, i.e. no exact location of objects in the image is provided. In general, PPR-FCN aims to calculate a score for each specific predicate associated with a pair of subject-object. PPR-FCN consists of two modules: WSOD g(weakly supervised object detection) and WSPP (weakly supervised predicate prediction). Since no bounding box of objects is given, the model does not learn to find objects and their locations in the image but makes a prediction based on the class of objects and their relative spatial relation. This strategy works well with semantic relationships but not occlusion/distance relationships because the occlusion relationship can be flipped even when two objects do not change their positions. Thus, compared to PPR-FCN, our model focuses on the position of objects and tries to attend to relevant areas only, which is a more local approach.}
 
\section{Conclusion}
We propose a novel three-decoder architecture as infrastructure and use GIT to enforce focused attention. Visualizations of attention weights in our occlusion decoder confirm the effectiveness of GIT in creating focused attention. Our proposed occlusion decoder that exhibits focused attention to task-relevant regions is more interpretable than decoders that attend to a large portion of the image. Like humans, our proposed occlusion decoder pays intense attention to relevant information and suppresses distracting information. Moreover, visualizations of attention weights and error analyses indicate that our proposed occlusion decoder is more robust, especially when dealing with mutual occlusions and objects very close to each other, further highlighting the importance of focused attention.

\section*{Acknowledgment}
The authors would like to express their sincere gratitude for Sijie Cheng's helpful discussions on attention heads. The authors would like to thank Jinjun Peng and Beiwen Tian for their help with GPU resources. The authors also thank Yupeng Zheng for his help with dataset processing.

\ifCLASSOPTIONcaptionsoff
  \newpage
\fi

\printbibliography

%

\begin{IEEEbiography}[{\includegraphics[width=1in,height=1.25in,clip,keepaspectratio]{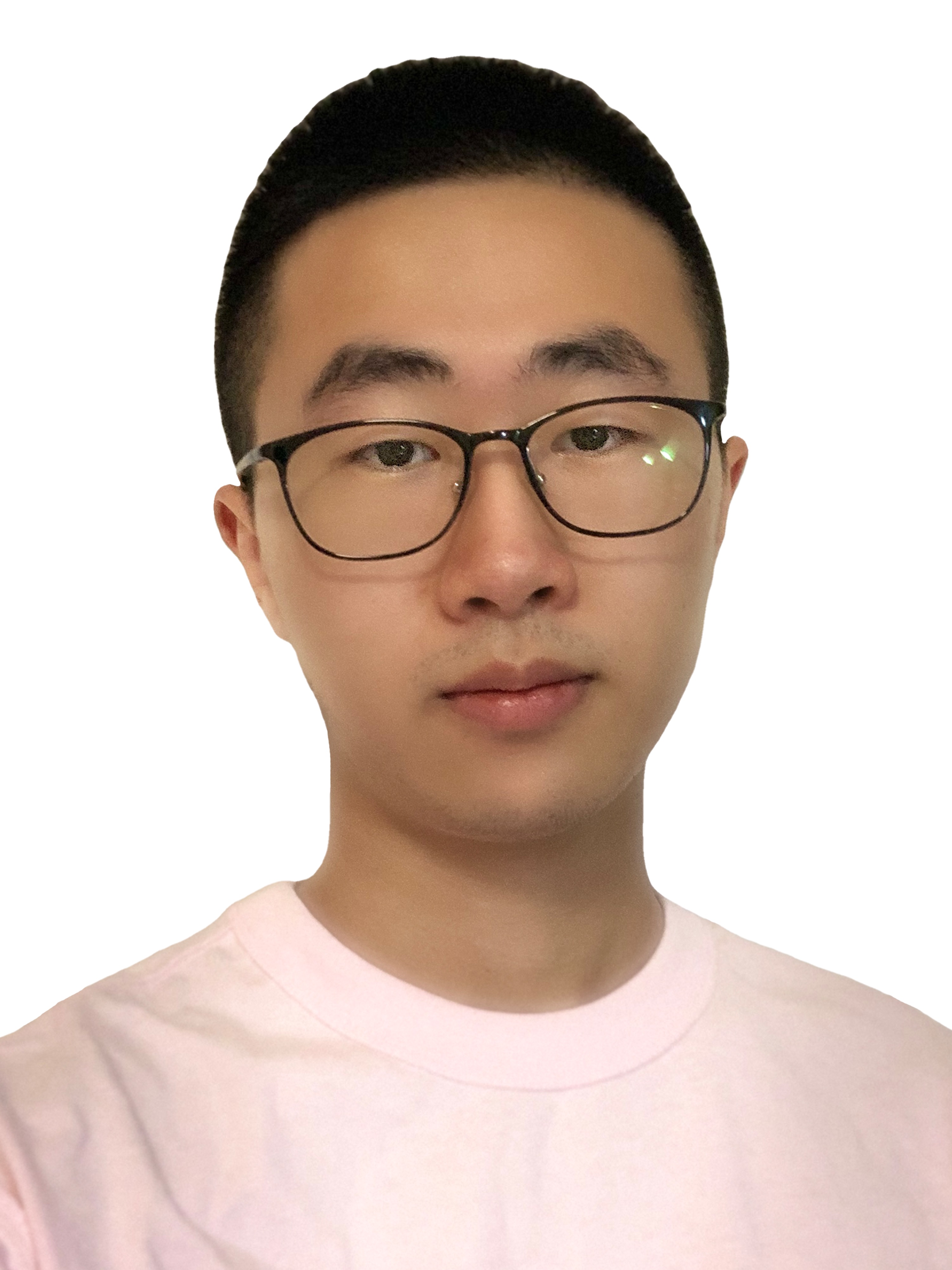}}]{Yang Li} received the B.S. degree in cognitive science with specializations in machine learning and neural computation from the University of California San Diego, La Jolla, CA, USA, in 2021. 

He is currently a research intern at Tsinghua University, Beijing, China. 
\end{IEEEbiography}

\begin{IEEEbiography}[{\includegraphics[width=1in,height=1.25in,clip,keepaspectratio]{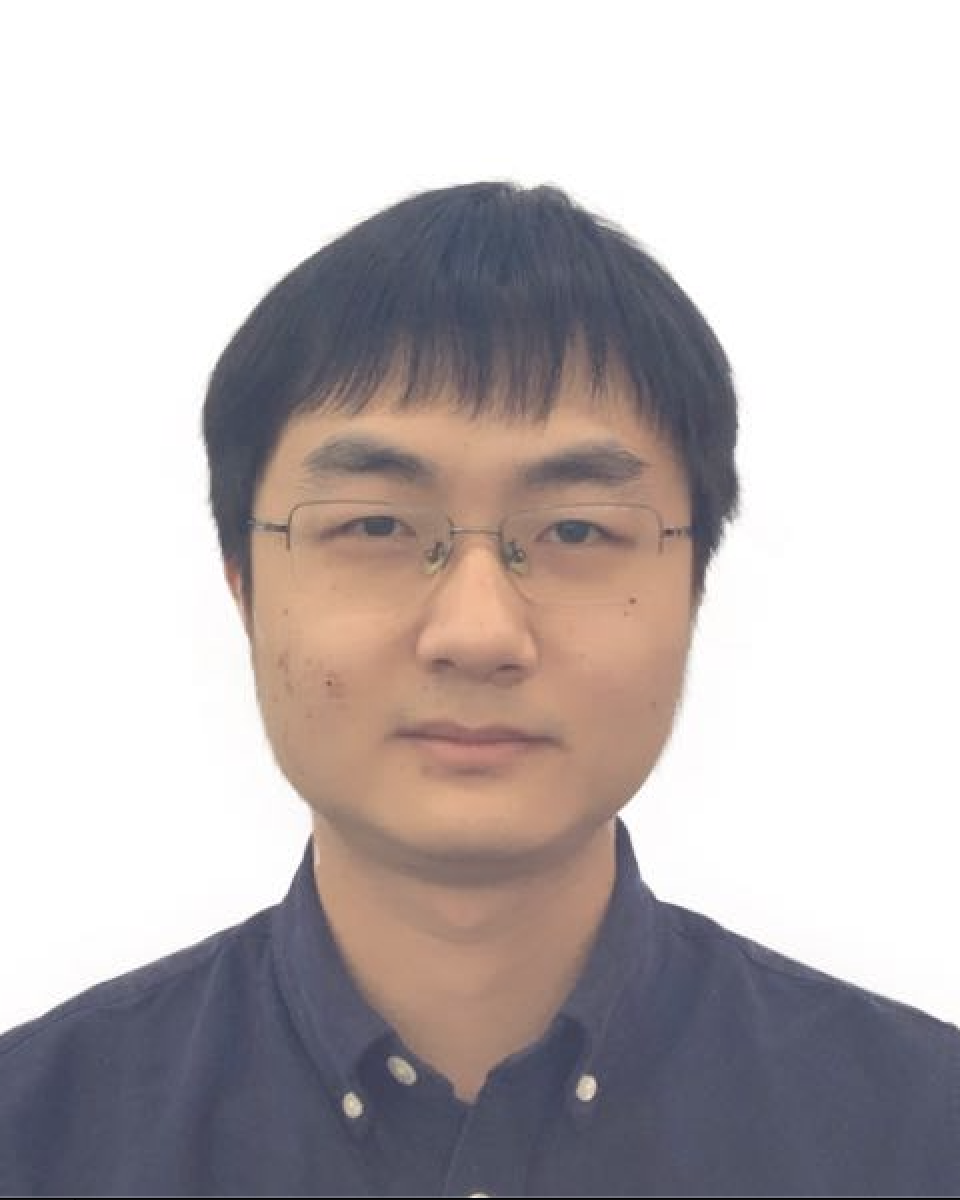}}]{Yucheng Tu}
obtained the B.S. degree in mathematics from Tsinghua University, Beijing, China, in 2015, and the Ph.D. degree in mathematics from the University of California San Diego, La Jolla, CA, USA, in 2021. His specialization is differential geometry and geometric analysis.

He is currently a teaching visitor at the University of California San Diego. Besides teaching, he is interested in research on deep learning and computer vision.
\end{IEEEbiography}

\begin{IEEEbiography}[{\includegraphics[width=1in,height=1.25in,clip,keepaspectratio]{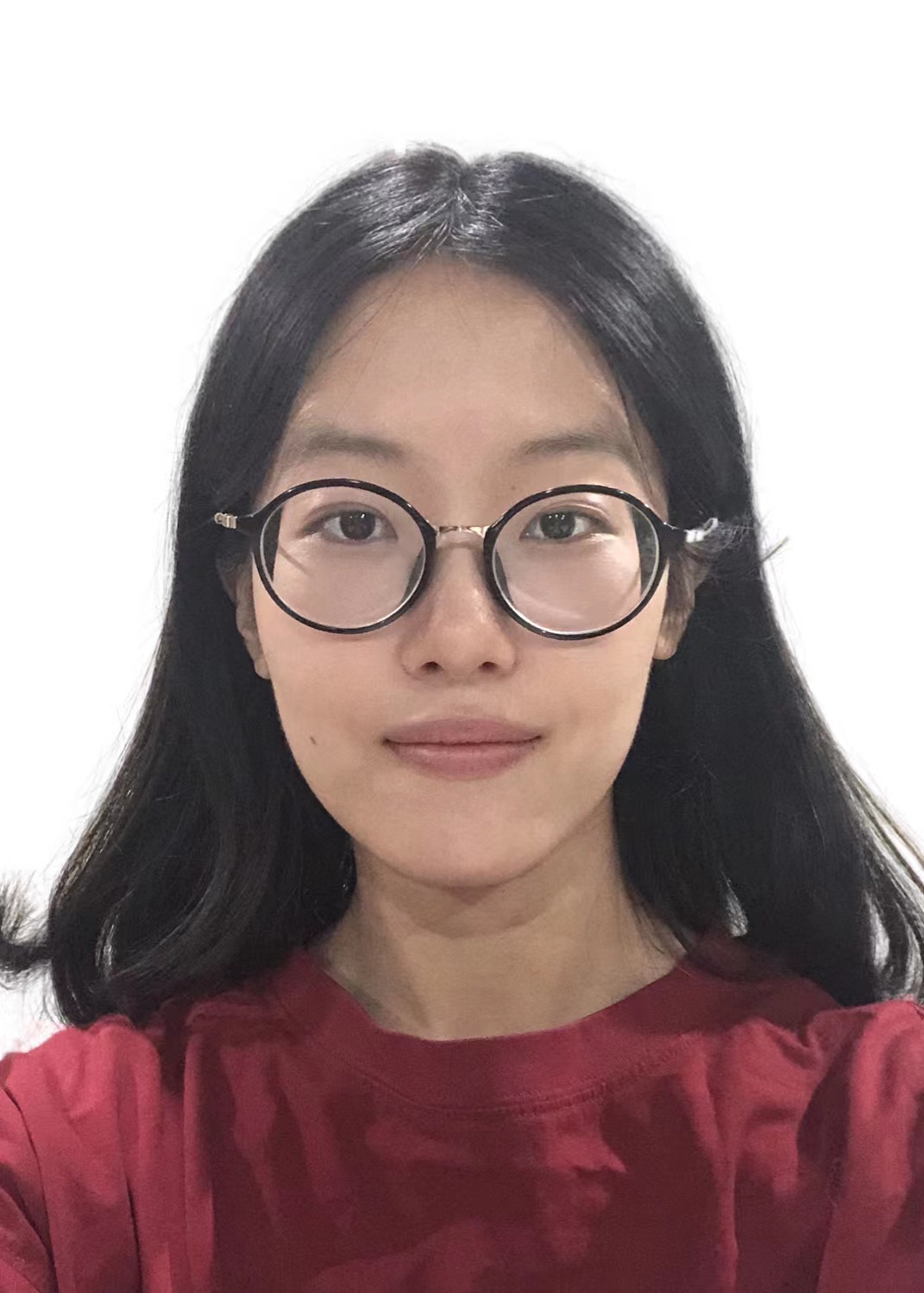}}]{Xiaoxue Chen}
obtained the B.S. degree in computer science and technology from Tsinghua University, Beijing, China, in 2021. She is currently a first-year Ph.D. student at the Institute for AI Industry Research, Tsinghua University. Her research interests are in the area of computer vision, especially 3D scene understanding. 
\end{IEEEbiography}


\begin{IEEEbiographynophoto}{Hao Zhao}
received the B.E. degree and the Ph.D. degree both from the EE department of Tsinghua University, Beijing, China. He is now a research scientist affiliated with Intel Labs China. He is also a joint postdoc affiliated with Peking University. His research interests cover various computer vision topics related to robotics, especially 3D scene understanding. Photograph not available at the time of publication.
\end{IEEEbiographynophoto}

\begin{IEEEbiographynophoto}{Guyue Zhou}
received the B.E. degree from Harbin Institute of Technology, Harbin, China, in 2010, and the Ph.D. degree from the Hong Kong University of Science and Technology, Hong Kong, China, in 2014. He is currently an Associate Professor with the Institute for AI Industry Research (AIR), Tsinghua University. His research interests include advanced manufacturing, robotics, computer vision, and human-machine interaction. Photograph not available at the time of publication.
\end{IEEEbiographynophoto}






\end{document}